\documentclass[journal]{IEEEtran} %18ページまで，12 以上で料金発生
\usepackage{amsmath,amsfonts}
\usepackage{array}
\usepackage{graphicx}
\usepackage{url}
\usepackage[hidelinks]{hyperref}
\usepackage{cleveref}
\usepackage{arydshln}
\usepackage{csquotes}
\usepackage[subrefformat=parens]{subcaption}
\usepackage{multirow}
\usepackage{balance}
\usepackage{bm}
\usepackage{xcolor}
\usepackage{bm}

\newcommand{\bhline}[1]{\noalign{\hrule height #1}}   

\newcommand{\projectpage}{\href{https://kokieto.github.io/EgoNoiseSeparation}{\textit{project page}}}
\renewcommand{\baselinestretch}{1}
\crefname{equation}{Eq.}{Eqs.}% {環境名}{単数形}{複数形} \crefで引くときの表示
\crefname{figure}{Fig.}{Figs.}% {環境名}{単数形}{複数形} \crefで引くときの表示
\crefname{table}{Table}{Table}% {環境名}{単数形}{複数形} \crefで引くときの表示
\def\BibTeX{{\rm B\kern-.05em{\sc i\kern-.025em b}\kern-.08em T\kern-.1667em\lower.7ex\hbox{E}\kern-.125emX}}

\makeatletter
\let\oldthebibliography\thebibliography
\let\endoldthebibliography\endthebibliography
\renewenvironment{thebibliography}[1]{%
  \oldthebibliography{#1}%
  \setlength{\itemsep}{0.4\baselineskip}% % ここを好みで調整：0.4～0.7\baselineskip
  \setlength{\parsep}{0pt}%
  \setlength{\parskip}{0pt}%
}{\endoldthebibliography}\makeatother
\makeatletter
\let\orig@thebibliography\thebibliography
\let\endorig@thebibliography\endthebibliography
\renewenvironment{thebibliography}[1]{%
  \orig@thebibliography{#1}%
  \fontsize{9pt}{9pt}\selectfont % ← ここで文字サイズ，行間を指定，デフォルトは9pt
}{\endorig@thebibliography}
\makeatother
\begin{document}
\bstctlcite{NoUseDashForRepeatedNames}
\bstctlcite{LimiNumberofAuthors}

\title{
    Open-Set Ego-Noise Separation for Legged-Robot Audition \\
    via Annotation-Free Adaptation and Pretrained-Model Transfer
}
\author{
    Koki Shoda$^{\dagger}$,
    Jun Younes Louhi Kasahara$^{\dagger}$,
    Aoba Koyanagi$^{\dagger}$,
    Qi An$^{\dagger}$,
    and Atsushi Yamashita$^{\dagger}$ \\
\thanks{
    This work was supported in part by JSPS KAKENHI Grant Number JP26KJ0923 and the World-leading Innovative Graduate Study Program for Proactive Environmental Studies (WINGS-PES), The University of Tokyo.
}
\thanks{Corresponding author: Koki Shoda.}
\thanks{$^{\dagger}$ The University of Tokyo, 7-3-1 Hongo, Bunkyo-ku, Tokyo 113-8656, Japan.}
}

\maketitle
\markboth{}%IEEE ROBOTICS AND AUTOMATION LETTERS.}
{}%Shoda \MakeLowercase{et al.}: Learning to Hear While Walking: Adaptive Ego-Noise Separation for Legged Security Robots}

\begin{abstract}
This paper proposes an open-set ego-noise separation framework for legged-robot audition via annotation-free adaptation and pretrained-model transfer.
The framework removes robot-specific ego-noise while preserving environmental sounds whose classes are not specified in advance.
Acoustic sensing provides cues about a robot's surroundings beyond the visual field, but walking-induced ego-noise from footstep impacts, joint-backlash rattling, and motor noise severely contaminates the recordings. 
The framework first uses RecurGraph to select ego-noise-dominant clips from the unlabeled recordings by aggregating clip embeddings into an embedding centroid and propagating scores over an audio-embedding graph.
The selected clips are mixed with diverse environmental sounds from a large-scale sound-event dataset to provide paired mixture--target supervision for open-set separation.
Transfer-DiT then adapts a general-purpose zero-shot neural separator to achieve high-fidelity open-set ego-noise separation for the target robot.
Experiments with bipedal and quadrupedal robots show reliable clip selection and improvements in separation quality and downstream task performance over baseline separators.
These results demonstrate the feasibility of annotation-free adaptation without separately recorded ego-noise-only data or manual clip-level annotations.
\end{abstract}


\section{Introduction}
\begin{figure}[t]
    \begin{minipage}[t]{0.49\linewidth}
        \centering
        \includegraphics[width=0.925\linewidth]{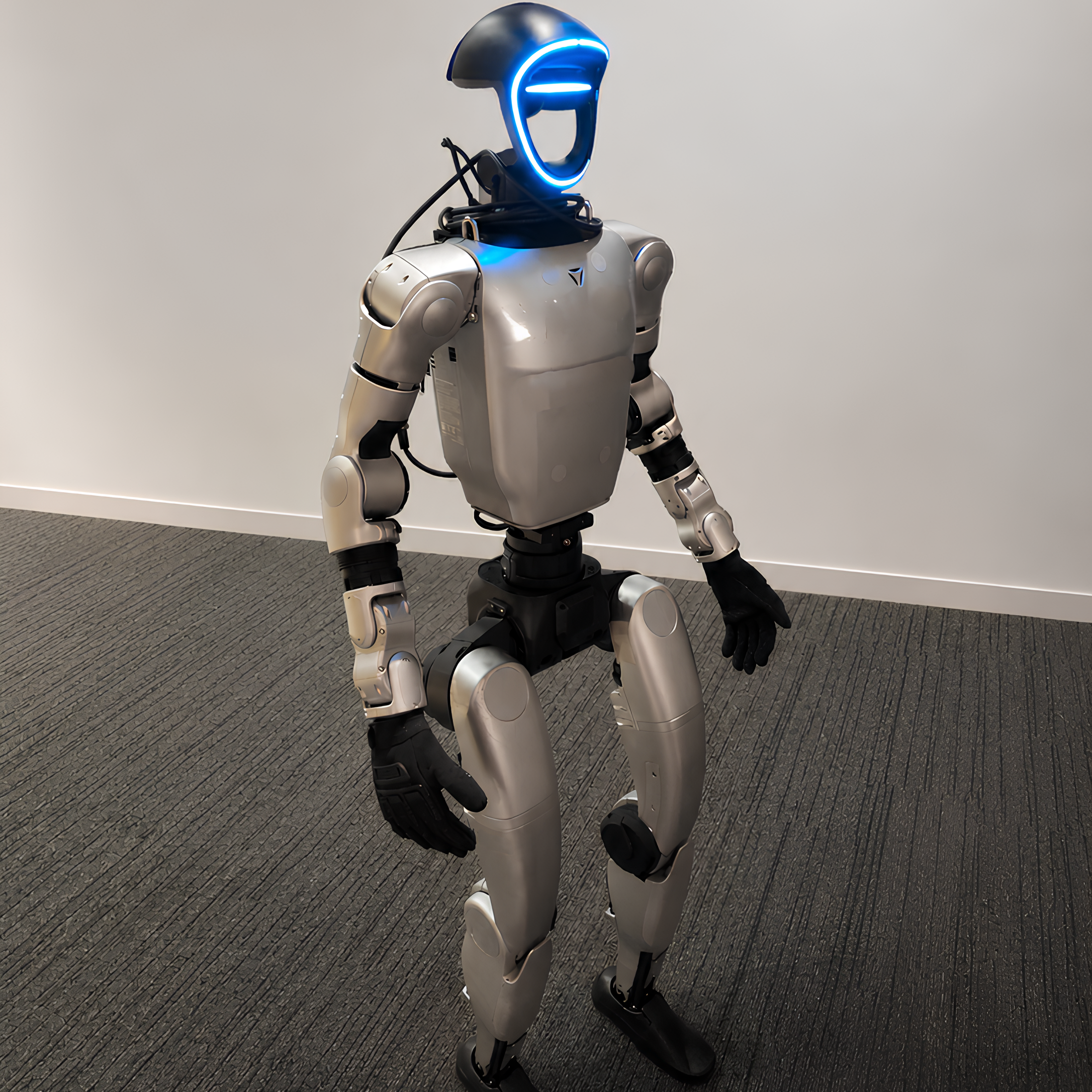}
        \subcaption{
            \footnotesize
            Bipedal robot.
        }
        \label{fig:robot-bipedal}
    \end{minipage}
    \hfill
    \begin{minipage}[t]{0.49\linewidth}
        \centering
        \includegraphics[width=0.925\linewidth]{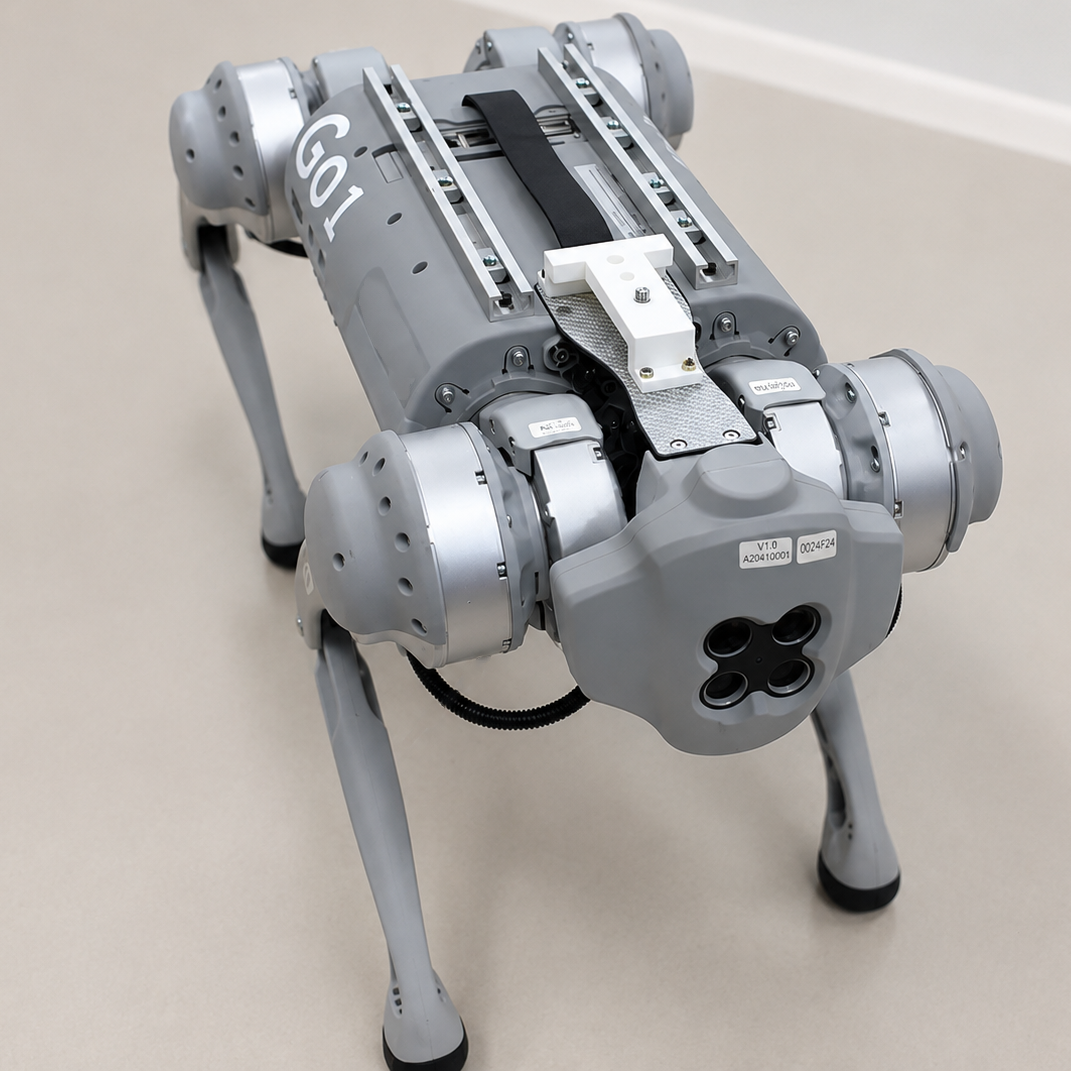}
        \subcaption{
            \footnotesize
            Quadrupedal robot.
        }
        \label{fig:robot-quadrupedal}
    \end{minipage}
    \caption{
      Legged robot examples.
    }
    \label{fig:robot}
\end{figure}

\IEEEPARstart{S}{ecurity} work, including patrol duties performed by security guards, often involves a substantial labor burden caused by night shifts and long working hours. 
Security robots that autonomously patrol indoor and outdoor facilities and detect anomalous events are therefore expected to play an important role~\cite{security_robot_survey, autonomous_patrol_robot}.
Legged robots, such as those shown in \cref{fig:robot}, can traverse steps and stairs that are difficult for wheeled robots~\cite{climing_robot}.
This mobility makes legged platforms especially promising for patrol applications.

Vision-based recognition has played a central role in robotic security~\cite{vision_suspicious_detection1, vision_suspicious_detection2}, but cameras are vulnerable to occlusion, darkness, and events outside the field of view.
Acoustic sensing holds great potential to complement vision.
Sound can propagate from behind obstacles and reveal security-relevant events such as gunshots, screams, and sirens~\cite{acoustic_anomaly_detection_security, anomalous_sound_detection_survey}.
However, onboard microphones also capture robot-generated sounds referred to as ego-noise~\cite{Shoda2024AR}.
For a walking robot, ego-noise includes footstep impacts, joint-backlash rattling caused by mechanical play in the joints, and motor noise.

Prior UAV studies have primarily focused on suppressing rotor noise to enhance speech~\cite{uav_ego_noise_reduction,uav_speech_enhancement}.
Security robots, however, must retain both speech and arbitrary environmental sound events whose classes cannot be specified in advance.
The task considered in this study is to remove robot-specific ego-noise while preserving arbitrary environmental sounds.
We refer to this task as \textit{open-set ego-noise separation}.

Legged-robot ego-noise depends on the platform and operating conditions.
High-fidelity suppression therefore requires a model that reflects the target robot and its intended operating domain.
Dictionary-based separators~\cite{SBINMF, MotorDictionaryNMF} can exploit robot-specific ego-noise characteristics, but representative separators require an ego-noise dictionary constructed from separately recorded ego-noise-only data.
Collecting ego-noise-only data separately for different robots, gaits, and operating conditions requires repeated preparation of quiet environments and imposes a substantial adaptation burden.

The objective of this study is to obtain robot- and domain-specific supervision from unlabeled pre-deployment recordings.
The recordings may contain both ego-noise-only clips and clips in which ego-noise overlaps with environmental sounds.
The pre-deployment recordings are intended to represent a target operating domain that can span multiple patrol sites and routes.
Even within a single site, changes in the robot's path and position alter foot--floor interactions and acoustic propagation paths.
The envisioned operational setting consists of two temporally and functionally distinct stages.
During \textit{pre-deployment adaptation}, the robot performs an adaptation walk, after which the recordings are processed and the separator is adapted offline.
During \textit{deployment}, the adapted parameters remain fixed and the separator processes newly acquired microphone signals online during routine patrol.

We realize the envisioned setting using a two-stage framework composed of our two proposed methods, RecurGraph and Transfer-DiT.
RecurGraph selects ego-noise-dominant clips from the unlabeled pre-deployment recordings by exploiting the recurrence of robot ego-noise across clips.
The selected clips are mixed with diverse environmental sounds sampled from a large-scale sound-event dataset, and the environmental sounds serve as training targets.
This training strategy exposes a neural separator to the broad empirical distribution of sounds to be preserved.
Transfer-DiT realizes this open-set training strategy with high fidelity by adapting a pretrained general-purpose neural separator to robot-specific ego-noise removal.

The contributions of this study are summarized as follows:
\begin{enumerate}
    \item \textbf{Open-set ego-noise separation:}
    We formulate open-set ego-noise separation based on the robot-specific ego-noise to be removed.
    We introduce a training strategy that mixes robot-specific ego-noise-dominant clips with diverse environmental sound clips sampled from a large-scale sound-event dataset.
    This strategy incorporates the broad empirical distribution of sounds to be preserved into paired mixture--target supervision, enabling a neural separator to learn open-set ego-noise separation.

    \item \textbf{Annotation-free adaptation:}
    We propose RecurGraph, which obtains robot-specific pseudo-supervision from unlabeled pre-deployment recordings through recurrence-guided seed initialization and graph-based score propagation, without separately recorded ego-noise-only data or manual clip-level annotations.

    \item \textbf{High-fidelity ego-noise separation:}
    We propose Transfer-DiT, which adapts a general-purpose pretrained Diffusion Transformer separator to robot-specific ego-noise removal.
    Transfer-DiT uses Bottleneck Adapter Layers~\cite{BAL} and Low-Rank Adaptation~\cite{lora} to preserve open-set environmental sounds with high fidelity.
\end{enumerate}

Videos and audio samples of the separation results are available on the \projectpage\footnote{Project page: \url{https://kokieto.github.io/EgoNoiseSeparation}}.
The RecurGraph and Transfer-DiT implementations are provided in the \href{https://github.com/kokieto/EgoNoiseSeparation}{\textit{repository}}\footnote{Repository: \url{https://github.com/kokieto/EgoNoiseSeparation}}, and the trained Transfer-DiT model weights are available on Hugging Face\footnote{Model weights: \url{https://huggingface.co/kokieto/EgoNoiseSeparation}}.

\section{Related Work}
\subsection{Platform-Dependent Characteristics of Ego-Noise}
Robot ego-noise depends strongly on the mechanical platform and its interaction with the environment.
UAV motors and propellers produce persistent ego-noise with prominent harmonic components~\cite{uav_ego_noise_reduction,uav_speech_enhancement}.
Wheeled robots and manipulators are commonly affected by motors, fans, drivetrains, and rolling mechanisms~\cite{mobile_robot_acoustic_signature,robot_ego_noise_template_subtraction}.
Legged robots additionally generate brief broadband footstep sounds through repeated foot--floor impacts~\cite{quiet_walking}.
Each impact excites structural floor vibration, and the resulting radiated sound depends on the mass, stiffness, damping, and internal structure of the contacted floor~\cite{footstep_floor_vibration,lightweight_floor_impact_sound}.
Reverberation and the positions and responses of the microphones further alter the recorded signal.
Consequently, legged-robot ego-noise varies across robots, gaits, floors, and recording conditions.
A fixed generic ego-noise model is inadequate for high-fidelity suppression.
The combination of broadband footstep impacts and floor-dependent sound radiation therefore poses modeling challenges that differ from persistent rotor- or motor-dominated ego-noise.

\subsection{Ego-Noise Reduction and Separation}
Source-level ego-noise reduction mitigates noise generation through passive mechanical treatments or motion-control policies.
For example, a quieter walking policy can be learned by incorporating footstep-noise reduction into a reinforcement-learning reward~\cite{quiet_walking}.
However, reducing foot impact introduces trade-offs with walking speed and energy consumption.
Source-level approaches can reduce footstep noise, although complete elimination of footstep noise and other actuator and contact sounds remains difficult.
Software-based separation is therefore still desirable.

Multichannel source-separation methods can exploit spatial information in addition to spectral and temporal features~\cite{ilrma,idlma,all_neural_beamforming,ArrayDPS,DroneEgoNoiseAutoencoder}.
On the other hand, the performance of multichannel separation strongly depends on the microphone-array configuration, the number of channels, geometric calibration, and reflections and occlusions caused by the robot body. 
In small legged robots and practical deployment environments, it may be difficult to secure a sufficient number of channels or an ideal microphone arrangement because of constraints on the available mounting space. 
More importantly, most multichannel methods internally rely on spectral information that can be obtained from a single channel. 
Therefore, improving single-channel separation performance remains an essential foundation even when extending the system to a multichannel setting. 
For these reasons, this study focuses on ego-noise separation from single-channel mixture recordings.

% Most multichannel methods internally rely on spectral information that can be obtained from a single channel.
% Therefore, improving single-channel separation performance remains an essential foundation even when extending the system to a multichannel setting.

Single-channel robot-specific approaches include semi-blind NMF and motor-data-driven NMF~\cite{SBINMF,MotorDictionaryNMF}.
These dictionary-based approaches can estimate the environmental-sound component by adapting its dictionary to each observed mixture and therefore do not require an environmental-sound training corpus~\cite{SBINMF}.
This flexibility is valuable for unknown target sounds, but low-rank additive spectral models have limited capacity to reconstruct complex environmental sounds when their time--frequency structures overlap with broadband and transient ego-noise.
Neural separators provide higher-capacity nonlinear mappings, but supervised training requires paired mixtures and target waveforms that define the signals to be preserved~\cite{uav_speech_enhancement}.
Prior neural-network-based robot ego-noise separation has primarily defined the preserved signal as speech~\cite{DeepClusteringEgoNoise, RotorConditionedDeepModels}.
To the best of our knowledge, prior neural-network-based robot ego-noise separation studies have not addressed this open-set setting.

% Realizing the open-set ego-noise separation with a neural separator requires training targets that represent the broad distribution of sounds occurring in real environments.

\subsection{Generic Audio Separation}
Large-scale pretrained audio-language models have enabled zero-shot source separation, in which a text prompt specifies a sound to extract or remove~\cite{audiosep,soloaudio}.
CLAPSep~\cite{clapsep} estimates a prompt-conditioned time--frequency mask.
SAM-Audio~\cite{sam_audio} uses a VAE~\cite{DAC-VAE} and Flow Matching~\cite{FlowMatching} to generate a separated latent representation.
These models provide broad audio priors and avoid task-specific training for every semantic source class.
Text prompts can describe semantic sound categories such as \enquote{mechanical footsteps}.
However, these prompts cannot sufficiently constrain the acoustic characteristics of ego-noise specific to the target robot.
Directly applying a generic zero-shot neural separator is therefore unlikely to achieve high-fidelity separation of robot-specific ego-noise.
The pretrained model nevertheless provides a useful foundation for robot-specific adaptation.

\begin{figure*}[t]
    \centering
    \includegraphics[width=0.9\linewidth]{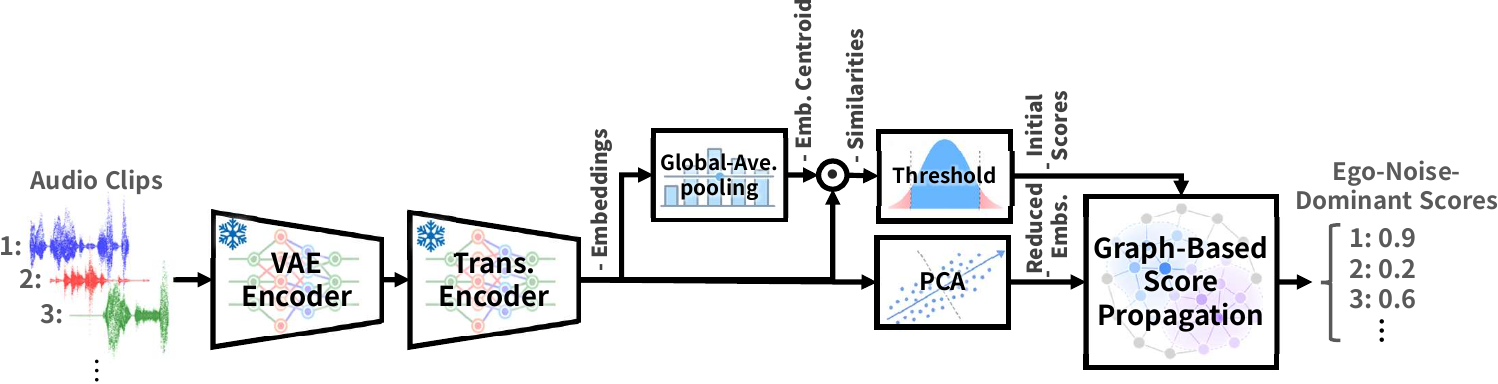}
    \caption{
        Overview of the proposed RecurGraph.
        The inputs are audio clips obtained by dividing recordings collected during pre-deployment adaptation walks into several-second segments.
        The VAE encoder and transformer encoder map each audio clip to an embedding.
        Global-average pooling of the embeddings constructs the embedding centroid, and similarities to the embedding centroid are thresholded to obtain initial scores.
        In parallel, Principal Component Analysis (PCA) produces reduced embeddings for graph construction.
        Graph-based score propagation propagates the initial scores over the reduced embeddings to obtain ego-noise-dominant scores for all audio clips.
    }
    \label{fig:recurgraph}
\end{figure*}

\section{Proposed Method}
\subsection{Concept}
The proposed framework embodies adaptive robot audition.
The robot learns its own ego-noise characteristics from pre-deployment recordings and uses the learned characteristics to improve environmental-sound perception.
Neural separators require paired mixture and target waveforms that represent both the robot-specific sound to be removed and the broad environmental-sound distribution to be preserved.
The proposed framework derives the robot-specific component from unlabeled pre-deployment recordings and samples preservation targets from a large-scale sound-event dataset.

RecurGraph exploits recurring ego-noise to select ego-noise-dominant clips from the unlabeled pre-deployment recordings.
Robot ego-noise recurs across clips recorded during walking, whereas environmental sounds are expected to occur intermittently and vary across clips.
RecurGraph exploits this occurrence-pattern difference to estimate an ego-noise-dominant score for each clip.
RecurGraph assumes that the adaptation recordings contain a sufficient proportion of ego-noise-dominant clips.
The selected clips are mixed with diverse environmental sounds, which serve as the training targets and expose the separator to the empirical distribution of sounds occurring in real environments.
Transfer-DiT then adapts a pretrained general-purpose separator to realize this open-set robot-specific removal function with high fidelity.

\subsection{RecurGraph: Recurrence-Guided Seed Initialization and Graph-Based Score Propagation}

The objective of RecurGraph is to automatically estimate an ego-noise-dominant score for each clip in the unlabeled pre-deployment recordings. 
In this study, ego-noise clip selection refers to identifying ego-noise-dominant clips in these recordings for use in separator training. 
Audio recorded in real environments contains robot ego-noise and environmental sounds that occur irregularly around the robot. 
If all recorded clips are used as ego-noise training data, the separator may incorrectly learn environmental sounds that should be preserved as ego-noise. 
To address this problem, RecurGraph combines recurrence-guided seed initialization with graph-based score propagation.
% RecurGraph does not require the input clips to contain only ego-noise.
% However, RecurGraph assumes that the adaptation recordings contain a sufficient proportion of ego-noise-dominant clips for recurring robot ego-noise to form the recording-wide common component.

Figure \ref{fig:recurgraph} illustrates the proposed procedure for estimating ego-noise-dominant scores. 
The input recordings are first divided into several-second audio clips. 
Distances computed directly in the raw waveform domain do not necessarily reflect the semantic similarity between clips. 
In particular, differences in phase, amplitude, and temporal alignment can produce large waveform distances even when the clips contain acoustically similar clips. 
We therefore use the audio branch of a pretrained audio-language embedding model~\cite{PE_AV}. 
The audio branch consists of a VAE encoder~\cite{DAC-VAE} followed by a Transformer encoder~\cite{attall}. 
The encoder maps each audio clip to an embedding in which semantically similar sounds are located close together.

Recording-level embedding aggregation is based on the difference in occurrence patterns between robot ego-noise and environmental sounds. 
Ego-noise, such as footstep impacts, joint-backlash rattling, and motor noise, is repeatedly generated while the robot walks and is therefore shared across a large number of clips. 
In contrast, environmental sounds are open-set and intermittent, and their acoustic characteristics vary across clips. 
We hypothesize that aggregation over a pre-deployment recording emphasizes acoustic components that recur throughout the recording.
Heterogeneous environmental sounds contribute less consistently to the aggregate representation. 
Based on this hypothesis, RecurGraph constructs the embedding centroid by applying global-average pooling over all clip embeddings.

The embedding centroid serves as a recording-derived reference embedding indicating the acoustic structure shared across the clip set. 
RecurGraph computes the cosine similarity between each clip embedding and the embedding centroid. 
A clip with high similarity is expected to be dominated by the recurring acoustic component, whereas a clip with low similarity is expected to contain acoustic characteristics that differ from this recording-wide component. 
The embedding centroid is constructed from the target recordings and can therefore reflect robot-, gait-, and recording-environment-specific characteristics.

The similarities to the embedding centroid are then converted into partially assigned initial ego-noise-dominant scores. 
Let $\rho$ denote the seed ratio, namely the proportion of clips assigned to each end of the similarity ranking.
The upper $\rho$ fraction of the clips in terms of similarity to the embedding centroid is assigned an initial ego-noise-dominant score of 1 (positive).
The lower $\rho$ fraction is assigned an initial score of 0 (negative). 
The initial scores of all remaining clips are left unassigned. 
Thus, the high-similarity clips serve as positive seeds, whereas the low-similarity clips serve as negative seeds for subsequent graph-based propagation.

RecurGraph uses relative ranking to assign the seed scores.
A fixed similarity threshold may not transfer across robots, gaits, and recording environments because the range and distribution of cosine similarities can change. 
In addition, the clips located in the middle of the similarity distribution may contain different relative levels of ego-noise and environmental sound. 
Directly assigning hard scores to these ambiguous clips could introduce incorrect seed assignments. 
RecurGraph therefore assigns initial scores only to the two ends of the similarity distribution.
The remaining clips stay unlabeled until their relationships with other clips are considered.

We hypothesize that ego-noise-dominant clips form locally coherent neighborhoods in the audio-embedding space because the clips share recurring robot-generated acoustic patterns. 
In contrast, clips containing environmental sounds tend to be more dispersed because of the diversity of environmental events. 
RecurGraph therefore uses similarity to the embedding centroid to define reliable positive and negative seeds.
The initial scores are then propagated through a neighborhood graph that captures the local structure of the audio-embedding space.
To capture these neighborhood relationships, RecurGraph constructs a graph from the audio embeddings. 
The latent representations obtained from the audio encoder may contain high-dimensional variations that are unnecessary for estimating ego-noise dominance.
Examples include subtle timbral differences and recording-condition-dependent components.
These variations can make local-neighborhood relationships sensitive to small, low-contribution embedding components.
Therefore, we apply PCA~\cite{PCA} before graph construction to obtain a compact representation of the dominant embedding structure.

Dimensionality reduction before extracting structure from neural embeddings is also an established pipeline in embedding-based text mining. 
Prior work has shown that clustering PCA-reduced pretrained word embeddings can recover coherent topics~\cite{embedding_topic_clustering}.
Systematic studies of document clustering have also examined pipelines combining BERT or Doc2Vec embeddings, PCA, and HDBSCAN~\cite{document_clustering_pipeline}. 
These studies provide precedent for exploiting the geometry of a reduced neural-embedding space to recover semantic group and neighborhood structure. 
Following this approach, RecurGraph uses PCA to construct a lower-dimensional space in which the relationships among the audio clips can be represented more compactly and stably.

PCA is applied only to the embeddings used for graph construction. 
The similarities to the embedding centroid and the corresponding initial scores are computed in the original pretrained embedding space so that the information encoded by the full audio representation is retained during score initialization. 
In parallel, the PCA-reduced embeddings are used to define the nodes and neighborhood relationships of the graph. Each node corresponds to one audio clip, and the graph connectivity represents acoustic similarity between clips in the reduced embedding space.
Finally, graph-based score propagation~\cite{local_global_consistency} assigns scores to the clips whose initial scores were left unassigned by propagating information from the positive and negative seeds. 
The graph therefore allows a clip with moderate similarity to the embedding centroid to receive a high ego-noise-dominant score when it is strongly associated with positive seeds through the local embedding structure. 
Conversely, a clip connected to negative seeds receives a low score even when its similarity to the embedding centroid alone is inconclusive.

RecurGraph combines the recording-wide common component represented by the embedding centroid with the local relationships represented by the graph to estimate an ego-noise-dominant score for every audio clip. 
Clips assigned high final scores are subsequently used to train the neural separator. 
The procedure requires neither a natural-language prompt, separately recorded ego-noise-only data, nor manual clip-level annotations.

\begin{figure*}[t]
    \centering
    \includegraphics[width=0.725\linewidth]{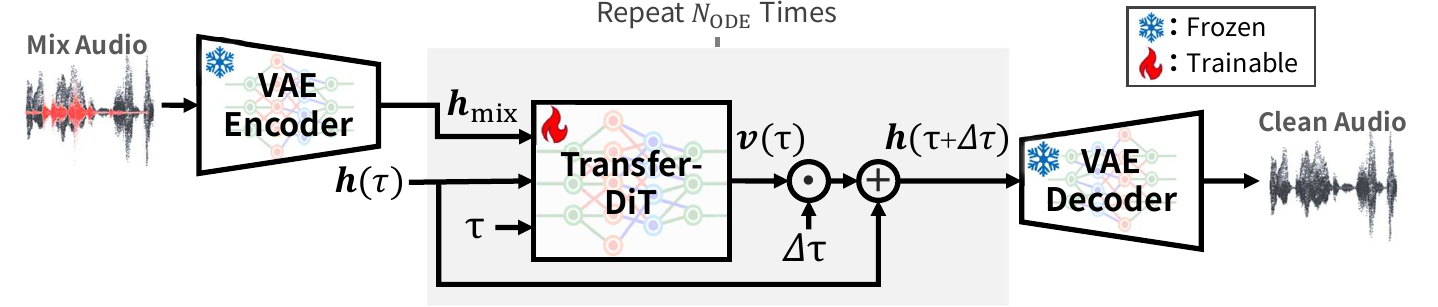}
    \caption{
        Overview of the latent-space Flow Matching procedure for ego-noise separation.
        A frozen VAE maps the mixture waveform to a latent representation.
        Transfer-DiT estimates the velocity field, and numerical integration produces an ego-noise-free latent that the frozen VAE decoder reconstructs as environmental sound.
    }
    \label{fig:over_arch}
\end{figure*}

\subsection{Open-Set Ego-Noise Separation}
\label{sec:training}
We formulate open-set ego-noise separation as a removal-defined problem.
The distribution of the source to be removed is known, whereas the signal to be preserved is not restricted to a predefined class.
Let the observed mixture $y(t)$ be
\begin{equation}
y(t)=s(t)+n(t), \qquad s(t)\sim p_{\mathrm{env}}, \quad n(t)\sim p_{\mathrm{ego}},
\end{equation}
where $t$ denotes time, $y(t)$ is the observed mixture waveform, $s(t)$ is the environmental-sound waveform, $n(t)$ is the ego-noise waveform generated by the target robot, $p_{\mathrm{env}}$ is the broad distribution of environmental sounds, and $p_{\mathrm{ego}}$ is the robot-specific ego-noise distribution.
Using the ego-noise-dominant clips selected by RecurGraph, we adapt the separator parameters $\theta$ to the robot-specific distribution $p_{\mathrm{ego}}$ and estimate
\begin{equation}
\hat{s}(t)=\mathcal{S}_{\theta}[y(t)],
\end{equation}
where $\mathcal{S}_{\theta}$ denotes the ego-noise separator parameterized by $\theta$, and $\hat{s}(t)$ is the estimated environmental-sound waveform.
Unlike speech enhancement, this formulation defines the separator by what should be removed rather than by what should be preserved.
This formulation is suitable for open-set ego-noise separation.
The separator must suppress only robot-specific ego-noise without removing unknown environmental sounds.

Dictionary-based separators~\cite{SBINMF,MotorDictionaryNMF} can estimate an environmental-sound component without training on an environmental-sound corpus, but their low-rank spectral representations restrict the fidelity of complex waveform reconstruction.
A supervised neural separator provides greater representational capacity but requires paired mixture and target waveforms.
We construct these pairs by mixing the ego-noise-dominant clips selected by RecurGraph with diverse environmental sound clips sampled from a large-scale sound-event dataset.
For an environmental sound $s(t)$ and a selected ego-noise-dominant clip $\tilde{n}(t)$, the training mixture is
\begin{equation}
y_{\mathrm{train}}(t)=s(t)+\tilde{n}(t),
\end{equation}
where $y_{\mathrm{train}}(t)$ is the training mixture waveform and $\tilde{n}(t)$ is the selected ego-noise-dominant waveform.
The corresponding environmental sound $s(t)$ is used as the training target.
A large-scale sound-event dataset provides an empirical approximation of $p_{\mathrm{env}}$ by capturing diverse temporal structures, spectral characteristics, and source types.
Pairing this empirical distribution with the selected ego-noise-dominant clips provides the paired mixture--target supervision needed to specialize a neural separator to $p_{\mathrm{ego}}$ while preserving open-set environmental sounds.

To improve robustness to different operating conditions, we apply data augmentations when constructing each mini-batch.
First, we randomly vary the mixing ratio between the ego-noise-dominant clip and the environmental sound clip within a specified Signal-to-Noise Ratio (SNR) range.
This augmentation exposes the separator to different relative levels of ego-noise and environmental sound.
Second, we convolve the clips with randomly selected Room Impulse Responses (RIRs).
This augmentation encourages robustness to diverse reverberation environments.
Third, we randomly perturb the playback speed of the ego-noise-dominant clips within a small range of a few percent.
This augmentation simulates slight changes in the pitch and rhythm of robot ego-noise caused by gait variations arising from floor inclination and payload.

\subsection{Transfer-DiT: Pretrained-Model Transfer for High-Fidelity Separation}
\label{sec:transfer_dit}
Transfer-DiT provides a high-fidelity deep-learning realization of the preceding open-set training strategy by adapting the pretrained DiT backbone of a general-purpose zero-shot neural separator~\cite{sam_audio} to robot-specific ego-noise separation.
Transfer-DiT uses conditional Flow Matching~\cite{FlowMatching} to learn a velocity field that transforms a Gaussian latent sequence into the environmental-sound latent sequence.
Figure~\ref{fig:over_arch} summarizes the resulting latent-space separation process.

Transfer-DiT performs separation in the VAE latent space.
This design exploits the general-purpose audio representations learned by the pretrained VAE and promotes generalization to environmental sounds unseen during pre-deployment adaptation.
The pretrained VAE encoder $E_{\mathrm{VAE}}$~\cite{DAC-VAE} maps the mixture waveform $y(t)$ and the corresponding environmental-sound waveform $s(t)$ to
\begin{align}
\bm{h}_{\mathrm{mix}} &= E_{\mathrm{VAE}}[y(t)], \\
\bm{h}_{\mathrm{env}} &= E_{\mathrm{VAE}}[s(t)],
\end{align}
where $\bm{h}_{\mathrm{mix}}$ and $\bm{h}_{\mathrm{env}}$ are the mixture and environmental-sound latent sequences, respectively.
The pretrained VAE architecture fixes the temporal sampling rate of both latent sequences at 25 Hz.
The same VAE encoder is used in RecurGraph, and all VAE parameters are frozen.

\begin{figure}[t]
    \centering
    \includegraphics[width=\linewidth]{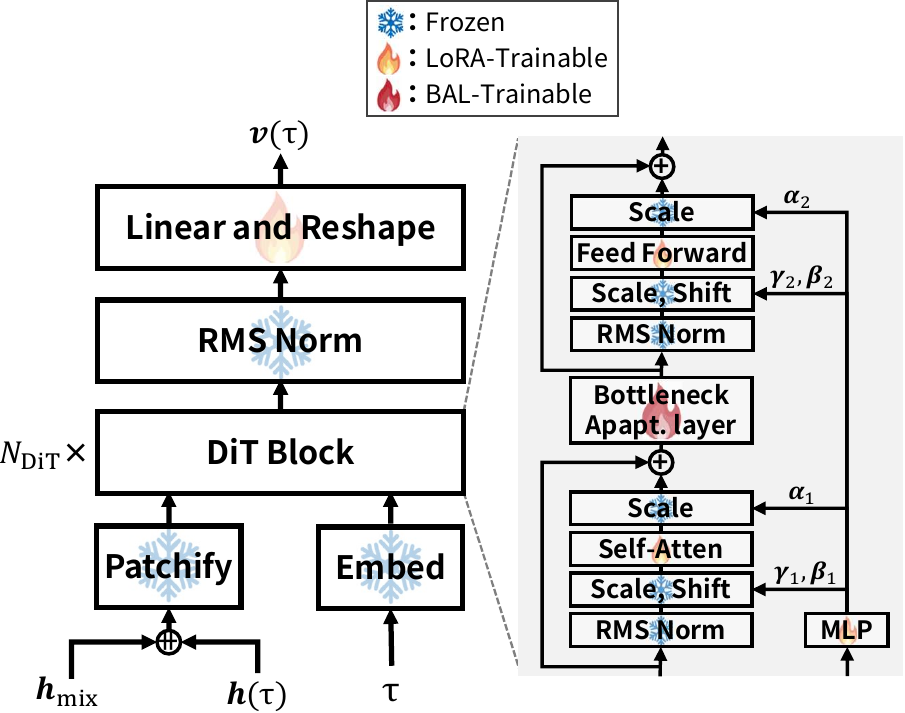}
    \caption{
        Transfer-DiT architecture.
        The pretrained DiT backbone of a general-purpose zero-shot neural separator is adapted to robot-specific ego-noise separation.
        The original target-conditioning modules, including the text encoder and cross-attention layers, are removed, while trainable BAL and LoRA provide robot-specific adaptation with the pretrained backbone fixed.
        This network architecture diagram follows the notation and bottom-to-top layout used in the original DiT paper~\cite{DiT}.
        }
    \label{fig:transfer-dit}
\end{figure}

Figure~\ref{fig:transfer-dit} details the Transfer-DiT architecture.
At flow time $\tau\in[0,1]$, $\bm{h}(\tau)$ denotes the current latent sequence.
The current latent $\bm{h}(\tau)$ is concatenated with the mixture latent $\bm{h}_{\mathrm{mix}}$ and patchified, whereas $\tau$ is embedded separately.
The resulting representations are processed by $N_{\mathrm{DiT}}$ DiT blocks.
After the final RMS normalization, the linear output head reshapes the representation into the conditional velocity $\bm{v}(\bm{h}(\tau),\tau,\bm{h}_{\mathrm{mix}})$.
Figures~\ref{fig:over_arch} and~\ref{fig:transfer-dit} abbreviate this conditional velocity as $\bm{v}(\tau)$.
At inference time, $\bm{h}(0)$ is initialized as a standard Gaussian latent sequence and numerically advanced to $\bm{h}(1)$ using the conditional velocity over $N_{\mathrm{ODE}}$ integration steps of size $\Delta\tau$, while $\bm{h}_{\mathrm{mix}}$ remains fixed as the condition.
The variables $N_{\mathrm{ODE}}$ and $\Delta\tau$ denote the number of integration steps and the integration step size, respectively, and $\bm{h}(\tau+\Delta\tau)$ denotes the updated latent sequence.
The frozen VAE decoder $D_{\mathrm{VAE}}$ reconstructs the environmental-sound estimate as $\hat{s}(t)=D_{\mathrm{VAE}}[\bm{h}(1)]$.

The original zero-shot neural separator~\cite{sam_audio} uses a text encoder and cross-attention layers to specify different target sources.
Transfer-DiT always removes robot ego-noise, so the target-conditioning modules are removed.
A trainable Bottleneck Adapter Layer (BAL)~\cite{BAL} is inserted at the original cross-attention residual position in every DiT block.
The pretrained backbone parameters remain frozen.
Low-Rank Adaptation (LoRA)~\cite{lora} updates the indicated linear projections, and the BAL parameters learn robot-specific residual transformations.
This adaptation strategy transfers the general source-separation capability of the pretrained backbone to the target robot and operating domain while keeping the backbone parameters fixed.

For training, we sample an initial latent $\boldsymbol{\epsilon}\sim\mathcal{N}(\bm{0},\bm{I})$ with the same dimensions as $\bm{h}_{\mathrm{env}}$.
We also sample a flow time $\tau\sim\mathcal{U}(0,1)$.
Here, $\bm{0}$ and $\bm{I}$ denote the zero vector and identity matrix, respectively.
The interpolated latent is $\bm{h}(\tau)=(1-\tau)\boldsymbol{\epsilon}+\tau\bm{h}_{\mathrm{env}}$, and the corresponding target velocity is $\bm{h}_{\mathrm{env}}-\boldsymbol{\epsilon}$.
The trainable parameters are optimized by minimizing
\begin{equation}
\mathcal{L}_{\mathrm{FM}}
=
\mathbb{E}
\left[
\left\|
\bm{v}
\left(\tau\right)
-
\left(
\bm{h}_{\mathrm{env}}
-
\boldsymbol{\epsilon}
\right)
\right\|^2
\right],
\end{equation}
where $\mathcal{L}_{\mathrm{FM}}$ denotes the conditional Flow-Matching loss.
The expectation $\mathbb{E}$ is taken over paired training latent sequences and the randomly sampled $\boldsymbol{\epsilon}$ and $\tau$.
The notation $\|\cdot\|$ denotes the Euclidean norm over all elements of a latent sequence.

\section{Experimental Setup}
\label{sec:exp_setup}
\subsection{Experimental Design Overview}

The experiments covered annotation-free pre-deployment adaptation and evaluation corresponding to the deployment stage.
The adapted parameters remained fixed during evaluation.
First, we constructed unlabeled pre-deployment recordings for each robot and evaluated whether RecurGraph identified ego-noise-dominant clips in the recordings.
Second, we used the selected ego-noise-dominant clips to adapt a pretrained separator to robot-specific ego-noise separation.
Third, we applied the adapted separator to reference-audio evaluation mixtures without further parameter updates.
The mixtures contained ego-noise recorded at sites disjoint from the adaptation sites and were used to evaluate robustness within the target operating domain.
This controlled protocol provided specified SNRs and isolated environmental-sound reference signals.
The protocol enabled consistent comparisons of source-separation quality and downstream Environmental Sound Classification (ESC) performance across methods.
Finally, we conducted a complementary physical playback evaluation.
The evaluation examined separator behavior when the mixture was acquired through the physical loudspeaker--room--microphone chain while the robot was walking.

For each robot, the unlabeled pre-deployment clips were the only robot-specific data provided to the proposed method during adaptation.
An ego-noise-only label indicated whether a clip contained no environmental sound.
We used the labels only to evaluate clip selection; the labels were not provided to the proposed method.

\subsection{Robots and Recording Conditions}

\begin{figure}[t]
    \begin{minipage}[t]{0.475\linewidth}
        \centering
        \includegraphics[width=0.95\linewidth]{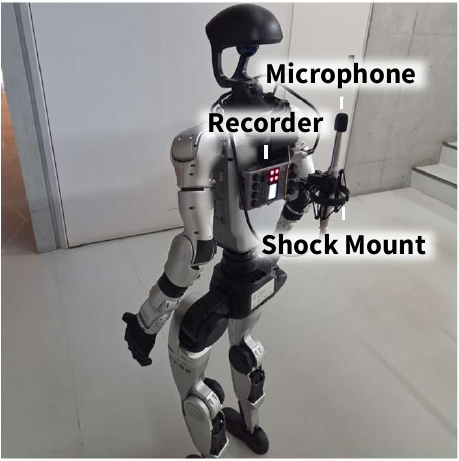}
        \subcaption{
            \footnotesize
            Recording hardware mounted on the Unitree G1 bipedal robot.
        }
        \label{fig:recording-hardware-bipedal}
    \end{minipage}
    \hfill
    \begin{minipage}[t]{0.475\linewidth}
        \centering
        \includegraphics[width=0.95\linewidth]{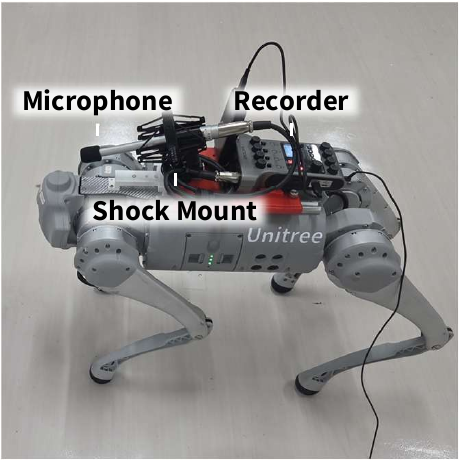}
        \subcaption{
            \footnotesize
            Recording hardware mounted on the Unitree Go1 quadrupedal robot.
        }
        \label{fig:recording-hardware-quadrupedal}
    \end{minipage}
    \caption{
        Robot configurations and recording hardware used in the experiments.
        % The microphone, recorder, and shock mount are labeled for each robot.
    }
    \label{fig:recording_hardware}
\end{figure}

We used the Unitree G1 bipedal robot and the Unitree Go1 quadrupedal robot.
The bipedal and quadrupedal configurations represent two common legged-robot morphologies.
Figure~\ref{fig:recording_hardware} shows the recording hardware mounted on each robot.
For both robots, the recording setup consisted of a microphone, recorder, and shock mount mounted on the robot body.
The shock mount reduced direct transmission of structure-borne vibrations from the robot body to the microphone, thereby reducing their contribution to the recorded signal.

\begin{figure}[t]   
    \begin{minipage}[t]{0.49\linewidth}
        \centering
        \includegraphics[width=0.95\linewidth]{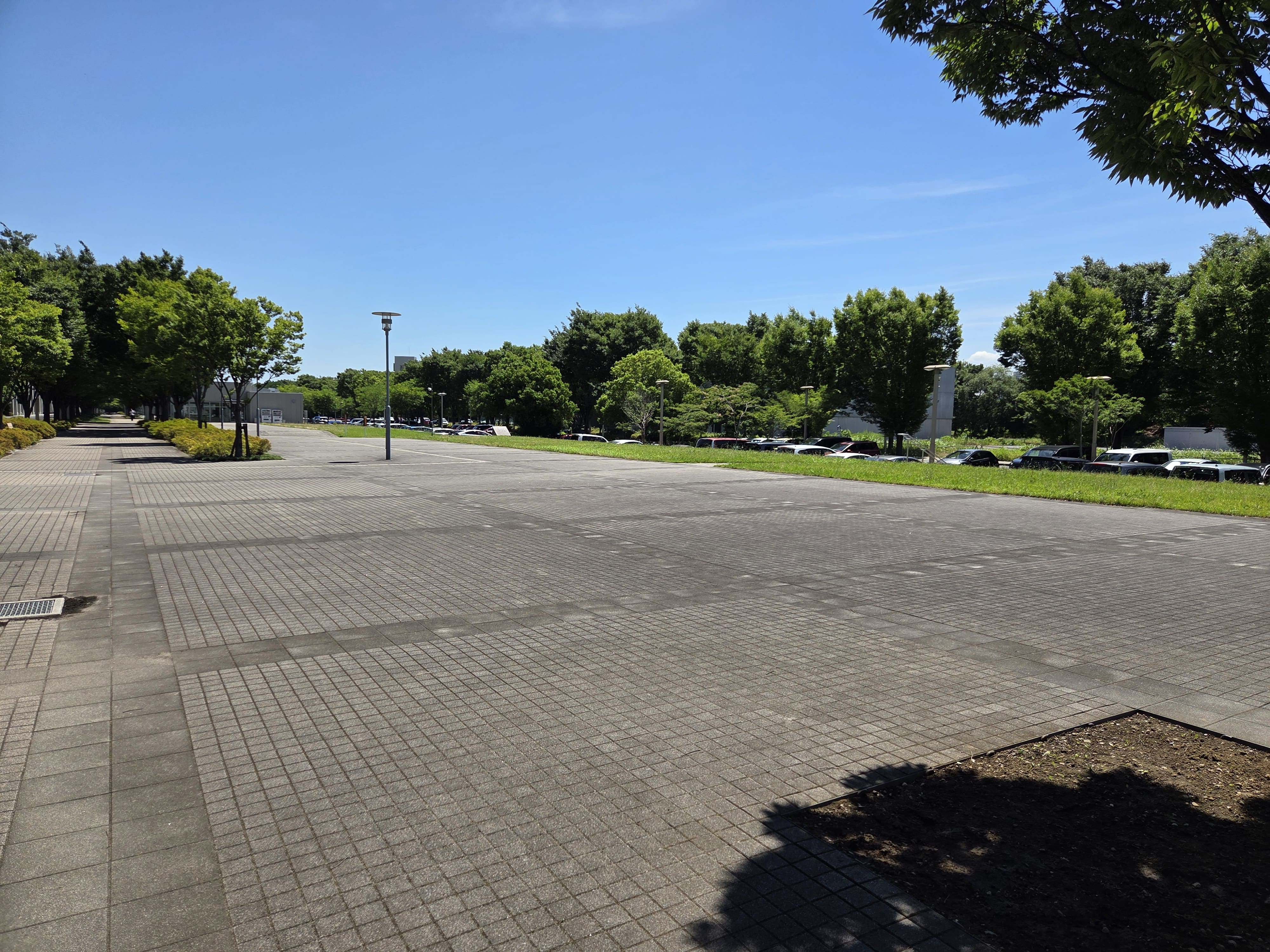}
        \subcaption{
            \footnotesize
            Outdoor plaza.
            % Floor material: concrete.
        }
        \label{fig:env-outdoor-plaza}
    \end{minipage}
    \hfill
    \begin{minipage}[t]{0.49\linewidth}
        \centering
        \includegraphics[width=0.95\linewidth]{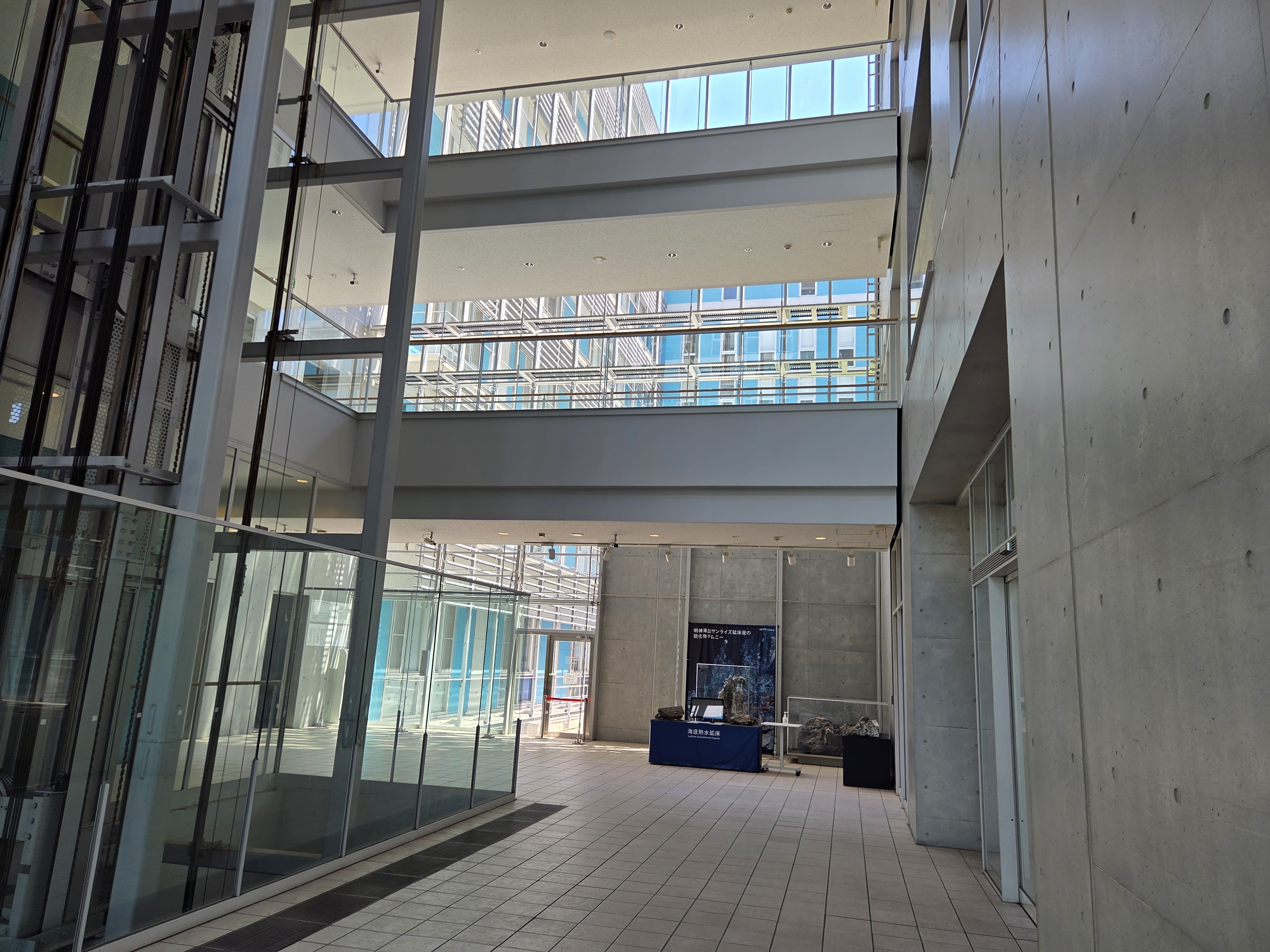}
        \subcaption{
            \footnotesize
            Indoor atrium.
            % Floor material: tile.
        }
        \label{fig:env-indoor-atrium}
    \end{minipage}
    \vspace{5pt}

    \begin{minipage}[t]{0.49\linewidth}
        \centering
        \includegraphics[width=0.95\linewidth]{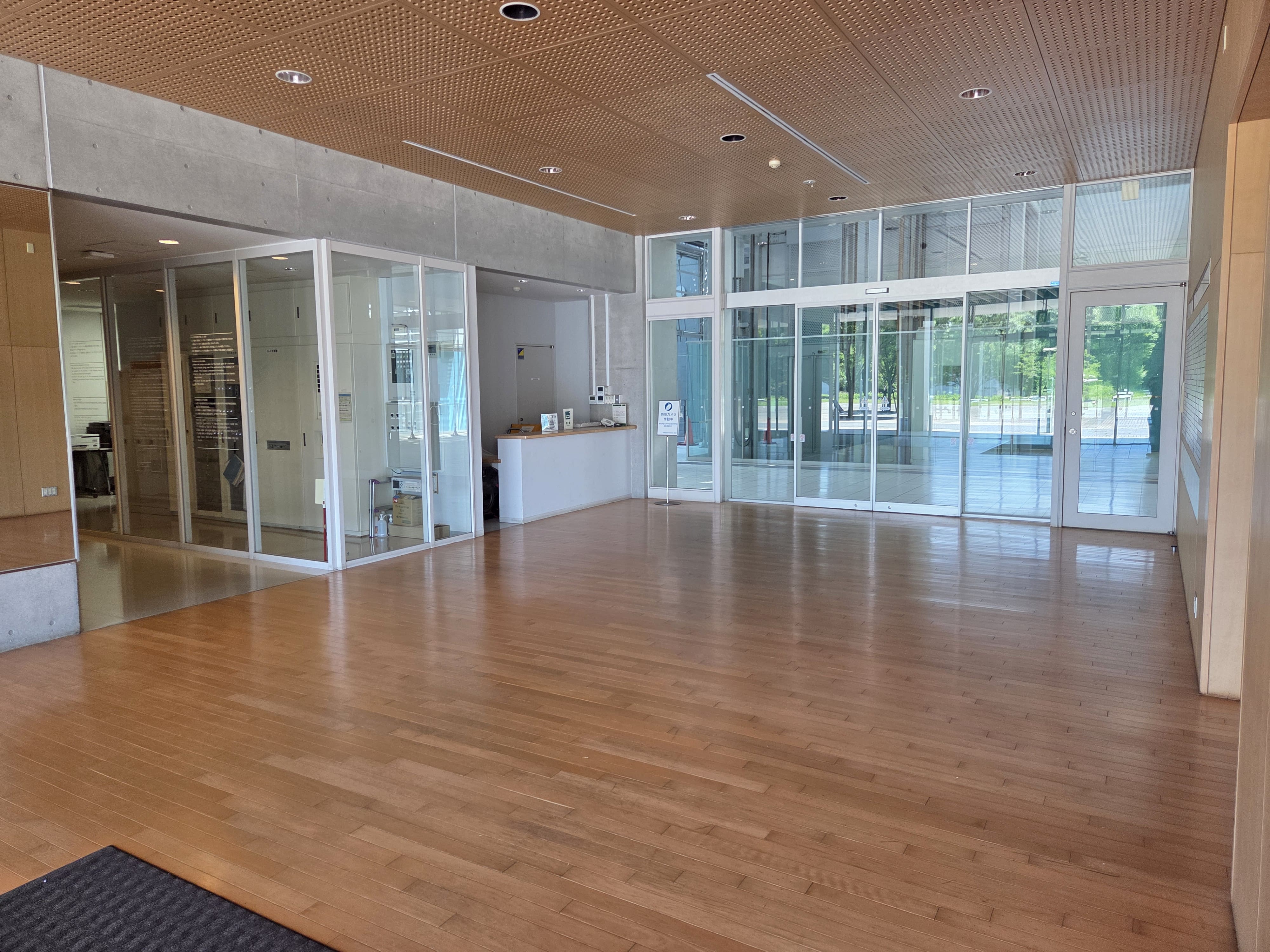}
        \subcaption{
            \footnotesize
            Building lobby.
            % Floor materials: wood and carpet.
        }
        \label{fig:env-entrance-lobby}
    \end{minipage}
    \hfill
    \begin{minipage}[t]{0.49\linewidth}
        \centering
        \includegraphics[width=0.95\linewidth]{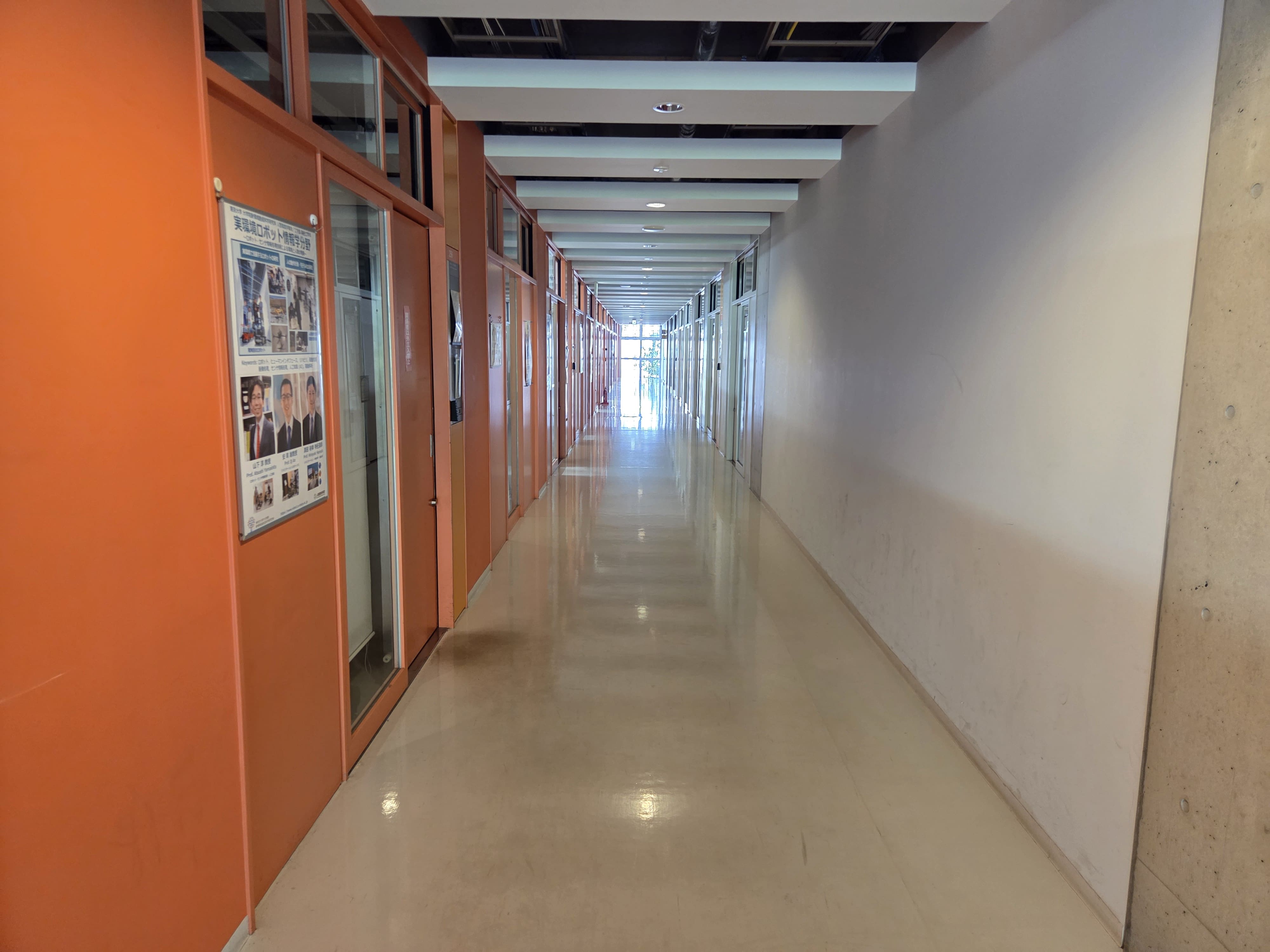}
        \subcaption{
            \footnotesize
            Corridor.
            % Floor material: vinyl.
        }
        \label{fig:env-corridor}
    \end{minipage}
    \caption{
        Environments in which ego-noise was recorded.
    }
    \label{fig:env}
\end{figure}

Figure~\ref{fig:env} shows representative environments in which the robot ego-noise was recorded.
All four recording sites were located within the same building and its immediate surroundings, which constituted the target operating domain in this study.
The sites comprised an outdoor plaza, an indoor atrium, a building lobby, and a corridor, spanning concrete, tile, wood, carpet, and vinyl floors.
Differences in surface stiffness and damping contribute to environment-dependent ego-noise through foot--floor interactions, together with reverberation and gait.
Variations in locomotion, including turning and changes in walking speed, were intentionally retained.
For both robots, recordings from the outdoor plaza and indoor atrium were used throughout pre-deployment adaptation, including the construction of separator-training data.
Separately recorded audio from the building lobby and corridor was held out for evaluation.
Room acoustics and floor conditions differed jointly between the adaptation and held-out sites.
The site-disjoint evaluation therefore measures robustness to combined shifts in room acoustics, floor conditions, and gait within the target operating domain.

The experimenter used separately recorded ego-noise-only signals to construct controlled mixture corpora and reference signals.
These signals were never provided to the proposed method as a separately identified clean ego-noise dataset.
Instead, the method received only the unlabeled pre-deployment recordings described in the following subsection.

\subsection{Adaptation Data and Ego-Noise Clip-Selection Evaluation}

The adaptation corpus simulated unlabeled pre-deployment recordings that would be obtained while each robot operated in the presence of open-set environmental sounds.
For each robot, we constructed 256 clips with a duration of 5~s from the adaptation-site ego-noise recordings.
Half of the clips contained only ego-noise.
The other half were generated by mixing ego-noise with environmental sounds sampled from the evaluation split of FSD50K~\cite{FSD50K}.
The SNRs were uniformly sampled from $-10.0$ to $10.0$~dB.
FSD50K contains 200 human-annotated sound-event classes and was used to provide diverse environmental sounds.
The clip-selection experiments used PE\_AV~\cite{PE_AV}, a state-of-the-art general-purpose audio--language embedding model, as the embedding backbone.
PE\_AV maps audio and text into a shared embedding space.
The same representation space therefore supports both the recording-derived reference embedding used by RecurGraph and a text-based prompt reference.
This property allowed us to compare similarity to the embedding centroid with similarity to a generic text-based prompt reference while keeping the embedding model fixed.
RecurGraph processed each robot's 256-clip set separately and received neither the ego-noise-only labels nor the isolated source signals.
The labels retained by the experimenter were used only to evaluate the resulting ego-noise-dominant scores.

We additionally evaluated the central assumption of RecurGraph under controlled changes in the composition of the adaptation recordings.
For each robot, the ego-noise-only ratio was selected from $\{0.25, 0.50, 0.75\}$ and the dominant environmental-class ratio was selected from $\{0, 0.25, 0.50, 0.75, 1.0\}$.
The latter ratio specifies the fraction of non-ego-noise-only clips whose environmental sound was drawn from one FSD50K primary class and therefore controls the prevalence of a frequently recurring environmental class.
For each robot and random seed, this class was sampled from primary classes containing at least 50 clips.
The remaining environmental sounds were drawn from other primary classes, and a ratio of zero did not enforce a frequently recurring environmental class.
Mixture SNRs were sampled from $-10$ to $10$~dB, and each condition was repeated over 20 random seeds.
We compared RecurGraph with direct PE\_AV similarity scoring using either the embedding centroid (embedding-centroid condition) or the robot-specific prompt reference (prompt-reference condition).
The evaluation metric was $\mathrm{AUPRC}_{\mathrm{ego}}$, where ego-noise-only clips were the positive class.

\subsection{Separator Training}

For each robot, clips whose RecurGraph score exceeded the selection threshold were selected for separator training.
The separator training data were generated on the fly by mixing the selected clips with environmental sounds from the training split of FSD50K~\cite{FSD50K} at SNRs uniformly sampled from $-12$ to $12$~dB.
The environmental sound was used as the training target, as described in \cref{sec:training}.
Twenty percent of the training pairs were held out for validation, and the checkpoint with the lowest validation Flow-Matching loss was used for evaluation.

The supervised neural separators were trained using the same selected clips and data-construction protocol as Transfer-DiT.
The ego-noise dictionaries for the dictionary-based separators were also constructed from the selected clips.
For SB-INMF~\cite{SBINMF}, we fixed the ego-noise dictionary and adapted an environmental-sound dictionary to each evaluation mixture using Itakura--Saito NMF.
For Harmonic NMF~\cite{MotorDictionaryNMF}, we constructed a time-varying harmonic dictionary from the oracle fundamental-frequency trajectory of the isolated ego-noise signal in each evaluation mixture.
For both NMF baselines, the dictionary rank was selected from $\{4, 8, 16, 32\}$ by maximizing the mean output SNR on held-out mixtures constructed from the selected clips and FSD50K training sounds.

To assess the effect of pretrained-model transfer, we also compared training the DiT from scratch with transfer learning using the same robot-specific training data.
For training from scratch, the DiT was randomly initialized and all its parameters were optimized.
For transfer learning, the pretrained backbone was used to initialize the model, and only the BAL and LoRA parameters were optimized.
Both strategies were trained for at most 200 epochs with a learning rate of $1.0 \times 10^{-4}$ and stopped when the validation loss did not improve for five consecutive epochs.
Their checkpoints were evaluated on the reference-audio mixtures at an SNR of 0~dB.

\subsection{Reference-Audio Evaluation}

The reference-audio evaluation used environmental sounds from 52 classes.
For the normal classes, we used 47 classes from ESC-50~\cite{ESC50}, excluding \enquote*{glass\_breaking}, \enquote*{siren}, and \enquote*{crying\_baby}.
For the anomalous classes, we used five classes in total: \enquote*{siren} from ESC-50~\cite{ESC50}, \enquote*{glass\_break}, \enquote*{gun\_shot}, and \enquote*{baby\_cry} from TUT Rare Sound Events 2017~\cite{TUTRareSoundEvents2017}, and \enquote*{screaming} from the Human Screaming Detection Dataset~\cite{HumanScreamingDetectionDataset}.
For each robot, we sampled 256 base environmental sounds, of which 26 (10\%) were anomalous and 230 (90\%) were normal.
Each base sound was mixed with ego-noise from the held-out recording sites at SNRs of $\{-10, -5, 0, 5, 10\}$~dB.
The same base sounds, ego-noise segments, and temporal offsets were shared across the five SNR conditions so that only the mixing ratio changed.

We compared Transfer-DiT with three groups of separators.
The zero-shot neural separators were CLAPSep~\cite{clapsep} and SAM-Audio~\cite{sam_audio}.
The supervised neural separators were Conv-TasNet~\cite{ConvTasNet}, Dual-Path RNN~\cite{DPRNN}, and Conformer~\cite{Conformer}.
The dictionary-based separators were SB-INMF~\cite{SBINMF} and Harmonic NMF~\cite{MotorDictionaryNMF}.
The unprocessed mixtures were also included as a reference condition.

Following prior work on Flow Matching-based source separation~\cite{sam_audio}, we used the audio-reference CLAP score, denoted by CLAP$_{\mathrm{A}}$~\cite{CLAPScore}.
CLAP$_{\mathrm{A}}$ evaluates the semantic similarity between the separated sound and the isolated reference environmental sound.
We used the SAJ score~\cite{SAJScore} to evaluate perceptual separation quality.
We also evaluated whether the separated environmental sounds support downstream classification using a simple training-free protocol.
The Area Under the Precision--Recall Curve, denoted by $\mathrm{AUPRC}_{\mathrm{anom}}$, measures binary discrimination between normal and anomalous sounds.
The 52-class accuracy, denoted by $\mathrm{Acc}_{52}$, measures classification performance among the 52 environmental-sound classes.
For this evaluation, we used PE\_AV~\cite{PE_AV} to compute the anomaly score as follows.

Let $\mathcal{N}$ and $\mathcal{A}$ denote the sets of normal and anomalous classes, respectively, and let $\mathcal{G}$ denote either class set.
For each class $c$, we used the text prompt \enquote{Audio of $c$}.
Given an input signal $x$, the audio embedding and text embedding for class $c$ are denoted by $\bm{z}_x$ and $\bm{z}_c$, respectively, and their cosine similarity is $s_c(x)=\bm{z}_x^\top\bm{z}_c$.
We computed the class-set score $S_{\mathcal{G}}(x)$ and anomaly score $p_{\mathrm{anom}}(x)$ as
\begin{gather}
S_{\mathcal{G}}(x)
=
\frac{1}{|\mathcal{G}|}
\sum_{c \in \mathcal{G}}
\exp\!\left(\frac{s_c(x)}{\alpha}\right),
\quad \mathcal{G}\in\{\mathcal{N}, \mathcal{A}\}, \\
p_{\mathrm{anom}}(x)
=
\frac{S_{\mathcal{A}}(x)}
{S_{\mathcal{N}}(x)+S_{\mathcal{A}}(x)}.
\end{gather}
Here, $|\mathcal{G}|$ denotes the number of classes in $\mathcal{G}$, and $\alpha=0.050$ is the temperature.
Averaging within each class set prevents the larger normal-class set from dominating the anomaly score.
The same prompts, temperature, and scoring rule were applied to every separation method and the unprocessed condition.
The scoring rule serves only as a fixed downstream evaluation protocol and is not a component of the proposed framework.

\begin{figure}[!t]
    \centering
    \includegraphics[width=0.7\linewidth]{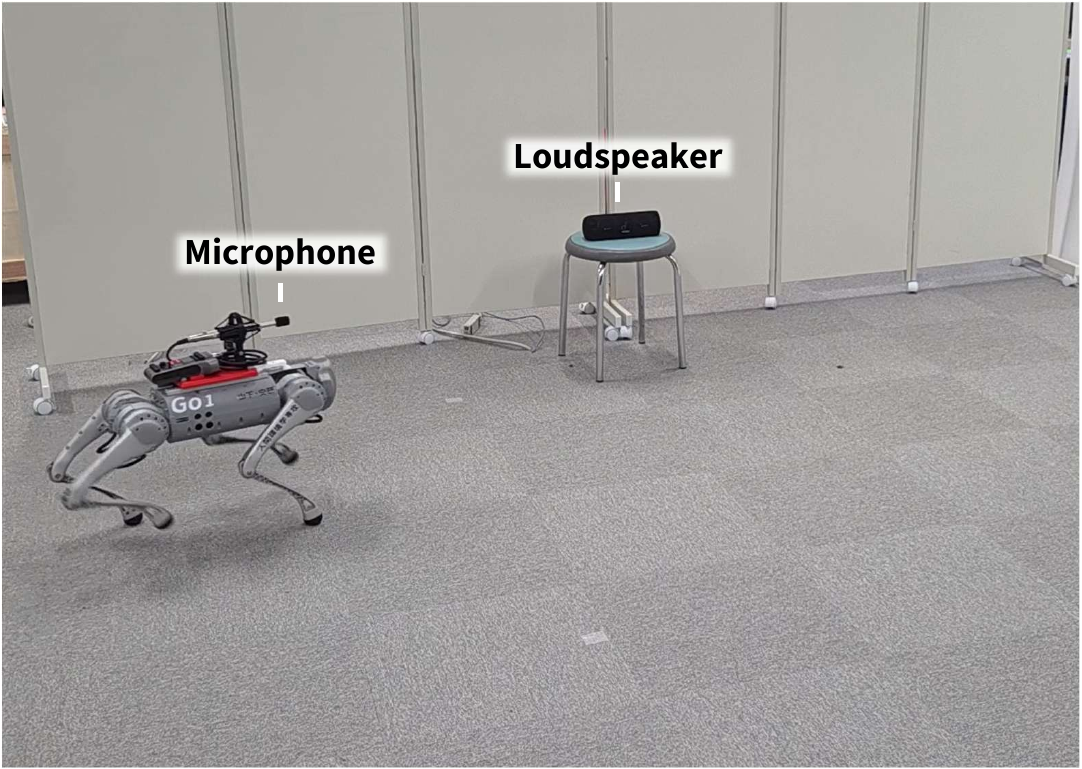}
    \caption{
        Physical playback experiment with the walking quadrupedal robot.
        Environmental sounds were played through a loudspeaker and recorded together with the robot's ego-noise.
    }
    \label{fig:physical_playback}
\end{figure}

\begin{figure*}[t]
    \centering
    \begin{minipage}[t]{0.38\linewidth}
        \centering
        \includegraphics[width=\linewidth]{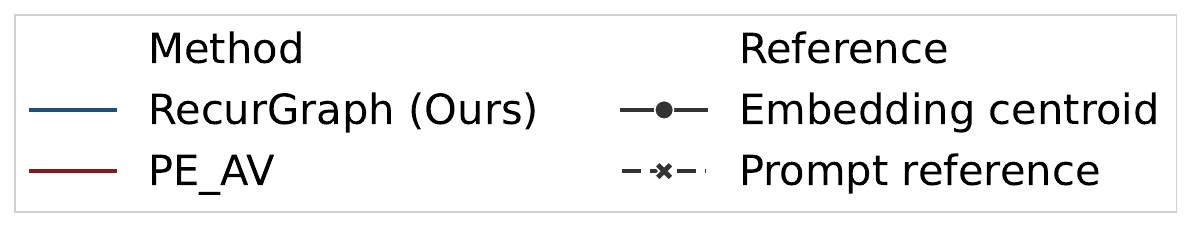}
    \end{minipage}
    \par\vspace{5pt}

    \begin{minipage}[t]{0.325\linewidth}
        \centering
        \includegraphics[width=\linewidth]{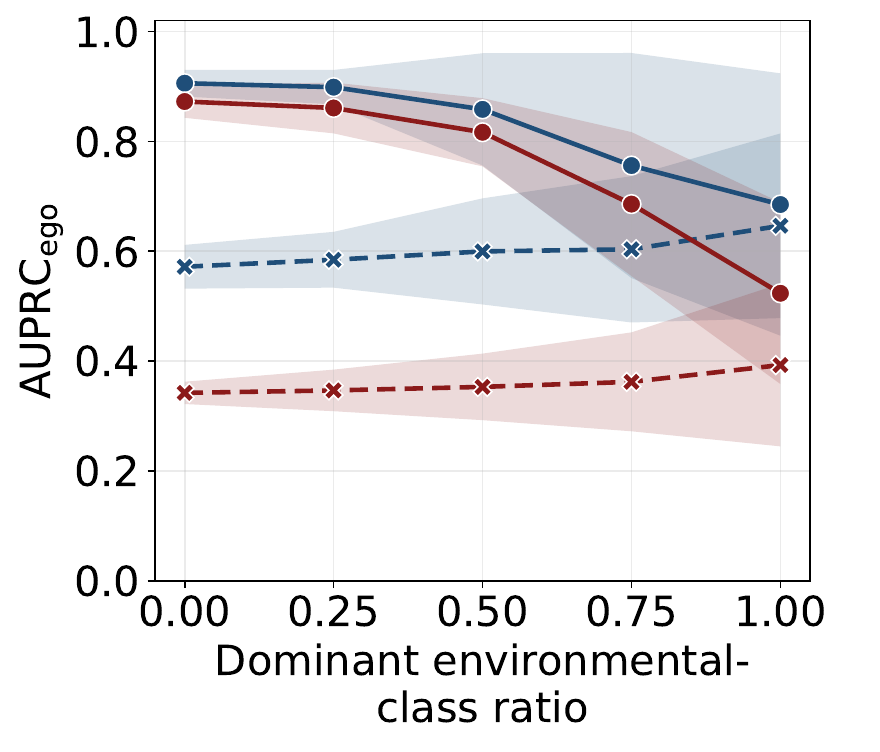}
        \subcaption{
            \footnotesize
            Ego-noise-only ratio $= 0.25$.
        }
        \label{fig:dominant_class_auprc_ego0p25}
    \end{minipage}
    \hfill
    \begin{minipage}[t]{0.325\linewidth}
        \centering
        \includegraphics[width=\linewidth]{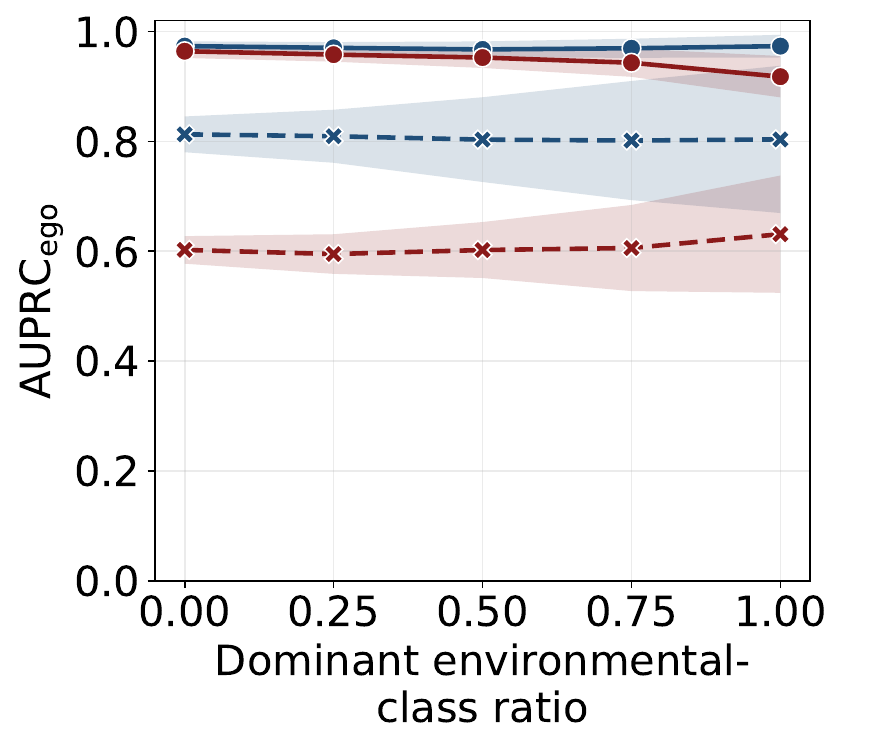}
        \subcaption{
            \footnotesize
            Ego-noise-only ratio $= 0.50$.
        }
        \label{fig:dominant_class_auprc_ego0p50}
    \end{minipage}
    \hfill
    \begin{minipage}[t]{0.325\linewidth}
        \centering
        \includegraphics[width=\linewidth]{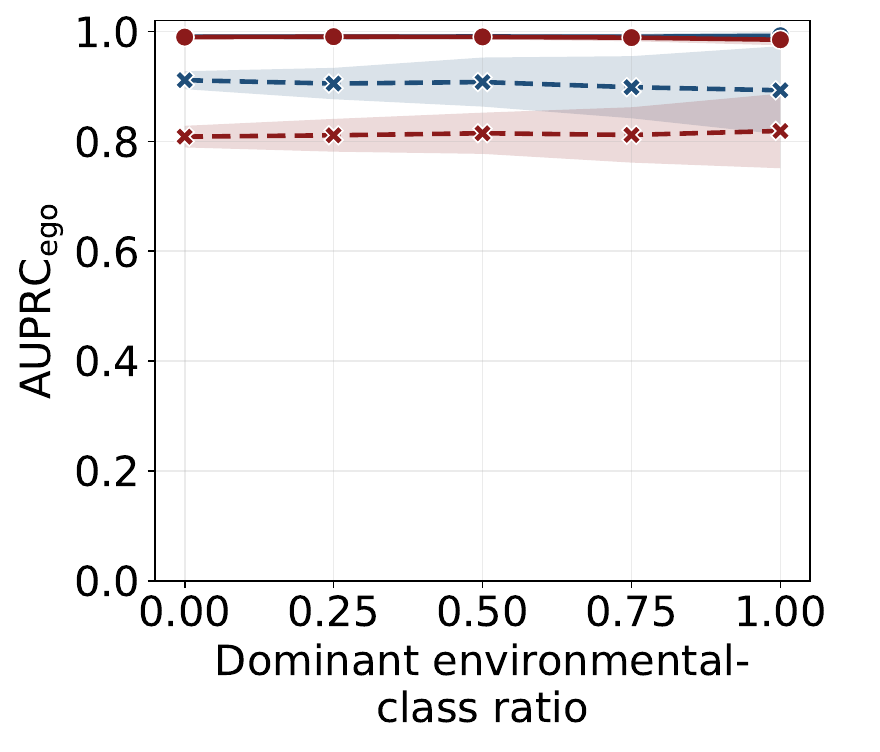}
        \subcaption{
            \footnotesize
            Ego-noise-only ratio $= 0.75$.
        }
        \label{fig:dominant_class_auprc_ego0p75}
    \end{minipage}
    \caption{
        Ego-noise clip-selection performance as the dominant environmental-class ratio varies at each fixed ego-noise-only ratio.
        RecurGraph is compared with direct PE\_AV similarity using the embedding centroid and prompt reference.
        Lines and shaded regions show the mean $\mathrm{AUPRC}_{\mathrm{ego}}$ and mean $\pm$ standard deviation over 20 random seeds, respectively.
    }
    \label{fig:dominant_class_auprc}
\end{figure*}

\subsection{Physical Playback Evaluation}
\label{sec:physical_playback_setup}
The controlled reference-audio evaluation provides specified SNRs and isolated environmental-sound reference signals for systematic comparisons across robots and separation methods.
To complement this controlled analysis, we conducted a physical playback experiment.
The mixture was acquired through the physical loudspeaker--room--microphone chain while the robot was walking.
The experiment evaluates separator behavior when the source-specific transfer paths and recording conditions are physically realized.
The physical playback site differed from the reference-audio evaluation sites and was not used to construct the training mixtures.
All separators were applied without further robot-specific adaptation to the physical playback recording.

Environmental sounds were played through a loudspeaker while the quadrupedal robot walked.
A microphone simultaneously captured the playback sounds and the robot's ego-noise, as shown in \cref{fig:physical_playback}.
We recorded 14 clips comprising six anomalous and eight normal playback events.
An operator manually controlled the robot using the joysticks on a handheld controller throughout the recordings.
The robot's walking speed, position, and heading differed across the 14 clips.
Each clip also contained variations in walking speed and heading as the robot moved and turned.
The clips were used to evaluate the unprocessed recordings and all applicable separators used in the preceding comparison.
Playable videos of this experiment with either the original recorded audio or the audio after ego-noise separation are available on the \projectpage.

Because the environmental sound and ego-noise were recorded only as a mixture, an isolated environmental sound was unavailable.
Therefore, instead of the reference-based CLAP$_{\mathrm{A}}$, we used the audio--text CLAP score, denoted by CLAP$_{\mathrm{T}}$~\cite{CLAPScore}.
CLAP$_{\mathrm{T}}$ is defined as the cosine similarity between the CLAP audio embedding of a separated sound and the CLAP text embedding of its playback event label.
We also report the reference-free SAJ score~\cite{SAJScore}, computed from the unprocessed recording, separated sound, and event-label text.
Harmonic NMF was inapplicable because the isolated ego-noise required to obtain its oracle fundamental-frequency trajectory was unavailable in the physical playback recording.

\subsection{Implementation Details}

The reference-audio mixtures and reference waveforms were materialized at 16~kHz, and each model resampled the signals internally when required by its frontend.
For RecurGraph, the seed ratio was set to $\rho=0.1$, PCA reduced the embeddings to 128 dimensions, and graph-based score propagation used a 64-nearest-neighbor graph.
Clips with an ego-noise-dominant score greater than $0.9$ were selected for separator training.
RecurGraph itself used no text prompt.
For the prompt-conditioned baselines, we considered more than 100 candidate descriptions of ego-noise and used PE\_AV to measure the similarity between each candidate and the separately recorded ego-noise-only signals.
We selected \enquote{mechanical footsteps} for the bipedal robot and \enquote{rapid mechanical footsteps} for the quadrupedal robot because these prompts provided a favorable balance between high similarity and descriptive simplicity.

Transfer-DiT was trained for 20 epochs using an initial learning rate of $1.0\times10^{-3}$, with both the BAL and LoRA ranks set to 16.
The learning rate was adjusted using a cosine annealing scheduler.
The validation loss was the Flow-Matching loss used for training, without separation inference or CLAP-based model selection during validation.
The playback speed of the selected ego-noise clips was randomly perturbed from $-2.5\%$ to $+2.5\%$.
RIR augmentation was applied with probability $0.25$; when applied, an RIR was sampled with equal probability from BUT Reverb Database~\cite{BUTRIR} or OpenSLR~\cite{OpenSLR}.
At inference time, the ODE was solved by the midpoint method using 16 uniform steps.

\begin{figure*}[t]
    \begin{minipage}[t]{0.49\linewidth}
        \centering
        \includegraphics[width=\linewidth]{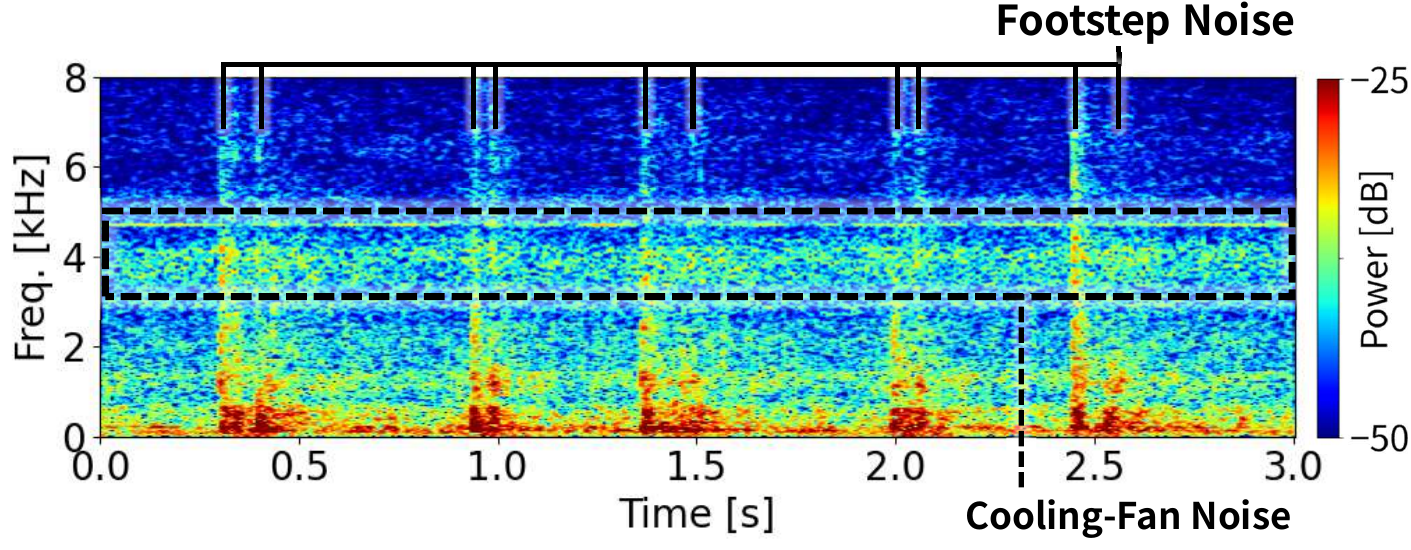}
        \subcaption{
            \footnotesize
            Ego-noise of the bipedal robot.
        }
        \label{fig:ego_noise_g1}
    \end{minipage}
    \hfill
    \begin{minipage}[t]{0.49\linewidth}
        \centering
        \includegraphics[width=\linewidth]{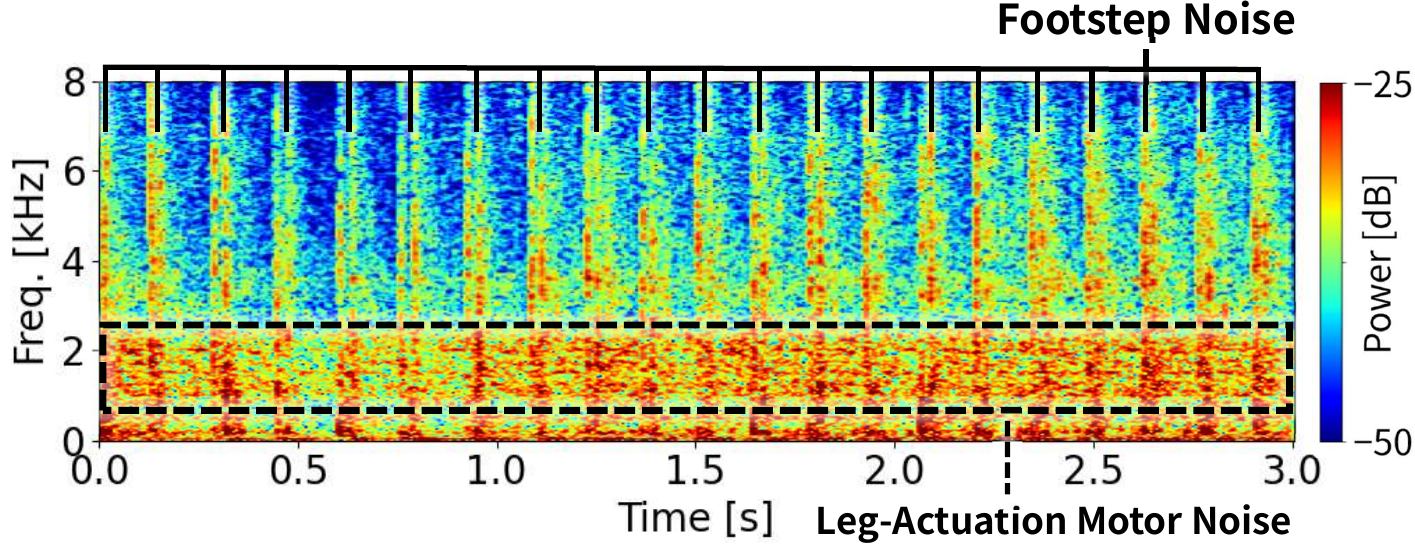}
        \subcaption{
            \footnotesize
            Ego-noise of the quadrupedal robot.
        }
        \label{fig:ego_noise_go1}
    \end{minipage}
    \caption{
        Representative spectrograms of robot ego-noise.
    }
    \label{fig:ego_noise_characteristics}
\end{figure*}

\section{Results and Discussion}

\subsection{Ego-Noise Clip Selection}

We evaluated whether RecurGraph can distinguish ego-noise-only clips from clips containing both ego-noise and environmental sound under changes in the composition of the unlabeled pre-deployment recordings.
The ego-noise-only ratio controls the prevalence of positive clips, while the dominant environmental-class ratio controls the prevalence of a frequently recurring environmental class among the negative clips.
The latter factor directly tests the premise underlying the embedding centroid: a frequently recurring environmental class can compete with recurring ego-noise in the recording-wide average embedding.

In \cref{fig:dominant_class_auprc}, increasing the dominant environmental-class ratio when the ego-noise-only ratio was $0.25$ reduced $\mathrm{AUPRC}_{\mathrm{ego}}$ for the embedding-centroid methods.
The reduction indicates that a frequently recurring environmental class makes the recording-wide common component less specific to ego-noise.
Nevertheless, RecurGraph using the embedding centroid remained more robust than direct PE\_AV similarity, particularly at high dominant environmental-class ratios.
RecurGraph also consistently outperformed direct PE\_AV similarity under the prompt-reference condition.
The improvement shows that graph propagation provides information beyond one-dimensional reference similarity.

As the ego-noise-only ratio increased, the embedding-centroid variants approached ceiling performance and became less sensitive to the prevalence of a frequently recurring environmental class.
This trend indicates that the embedding centroid is most reliable when ego-noise-only clips form a sufficiently large recurring component of the adaptation recordings.
The prompt-reference results remained lower than the corresponding embedding-centroid results, especially when ego-noise-only clips were sparse.
This result favors recording-specific acoustic recurrence over a generic text description.
Overall, the embedding centroid supplies recording-specific initial scores.
Graph propagation improves the classification of clips that cannot be reliably ranked by reference similarity alone.

RecurGraph outperformed direct PE\_AV similarity and was more robust to the presence of a frequently recurring environmental class.
However, applying RecurGraph requires a sufficiently large subset of ego-noise-dominant clips in the adaptation recordings.
The required proportion is not universal because the reliability of the recording-wide common component also depends on the prevalence of a frequently recurring environmental class.
Collecting adaptation recordings over diverse times, locations, and walking trajectories is expected to reduce the prevalence of a frequently recurring environmental class and increase the availability of ego-noise-dominant clips.

\begin{figure*}[t]
    \centering
    \includegraphics[width=0.7\linewidth]{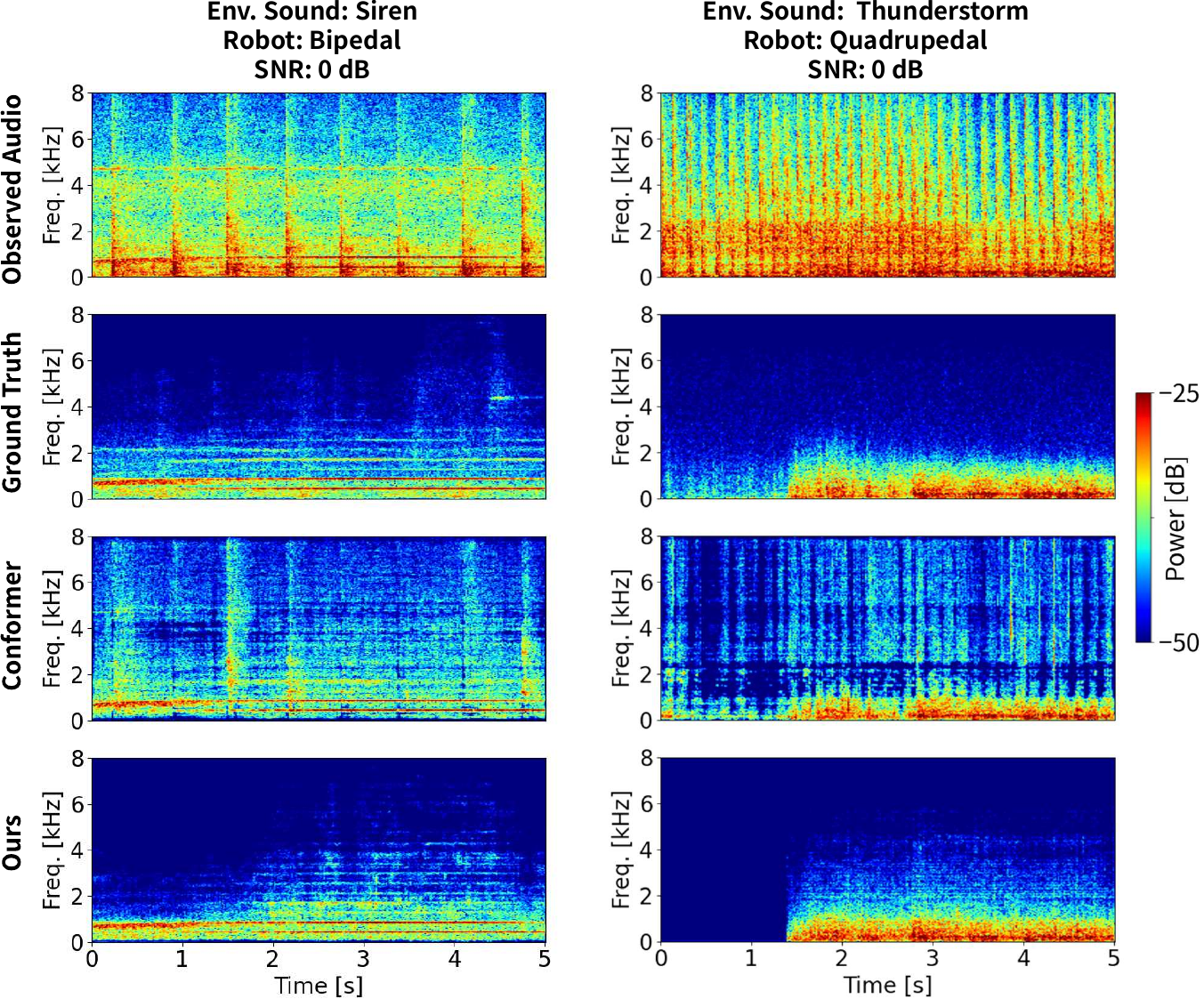}
    \caption{
        Comparison of ego-noise separation results of Conformer~\cite{Conformer} and our method for the bipedal robot (left) and quadrupedal robot (right).
        Additional comparisons with other methods and playable audio examples are available on the \projectpage.
    }
    \label{fig:examples}
\end{figure*}

\subsection{Ego-Noise Separation and ESC}
Figure~\ref{fig:ego_noise_characteristics} shows representative ego-noise spectrograms for the two robots.
The bipedal robot produced approximately $2.0$ footstep events per second and steady cooling-fan noise concentrated at approximately $3.0$--$5.0$~kHz.
Because the bipedal robot has ankle joints, each step produced two closely spaced impacts, which appeared as paired vertical structures in the spectrogram.
In contrast, the quadrupedal robot produced approximately $7.0$ footstep impacts per second together with leg-actuation motor noise concentrated at approximately $1.0$--$3.0$~kHz.
The bipedal ego-noise combined intermittent mechanical impacts with a stationary fan component, whereas the quadrupedal ego-noise was dominated by dense footstep impacts and motor actuation.
The contrasting time--frequency structures provide context for the separation and ESC results below.

\begin{table*}[t]
    \centering
    \caption{
        Comparison of DiT initialization and adaptation at an SNR of 0~dB.
        Stop epoch denotes the early-stopping epoch.
        CLAP$_{\mathrm{A}}$ and SAJ are means over 256 evaluation clips per robot, and $\mathrm{Acc}_{52}$ and $\mathrm{AUPRC}_{\mathrm{anom}}$ are computed over the same clips.
        The best value for each robot is shown in bold.
    }
    \label{tab:training_strategy_comparison}
    \footnotesize
    \setlength{\tabcolsep}{3pt}
    \begin{tabular}{l|ccccc|ccccc}
        \bhline{1pt}
        & \multicolumn{5}{c|}{Bipedal robot} & \multicolumn{5}{c}{Quadrupedal robot} \\
        DiT Configuration
        & \shortstack{Stop Epoch$\downarrow$}
        & CLAP$_{\mathrm{A}}\!\uparrow$
        & SAJ$\uparrow$
        & $\mathrm{Acc}_{52}\!\uparrow$
        & $\mathrm{AUPRC}_{\mathrm{anom}}\!\uparrow$
        & \shortstack{Stop Epoch$\downarrow$}
        & CLAP$_{\mathrm{A}}\!\uparrow$
        & SAJ$\uparrow$
        & $\mathrm{Acc}_{52}\!\uparrow$
        & $\mathrm{AUPRC}_{\mathrm{anom}}\!\uparrow$ \\
        \hline
        From Scratch
        & 137 & 0.53 & 3.1 & 0.52 & 0.77
        & 194 & 0.52 & 3.3 & 0.43 & 0.88 \\
        Pretrained Transfer
        & \textbf{29} & \textbf{0.73} & \textbf{3.8} & \textbf{0.73} & \textbf{0.94}
        & \textbf{55} & \textbf{0.71} & \textbf{3.8} & \textbf{0.70} & \textbf{0.93} \\
        \bhline{1pt}
    \end{tabular}
\end{table*}

\begin{table*}[t]
    \centering
    \caption{
        Separation performance on the evaluation data.
        The best value in each column is shown in bold.
    }
    \label{tab:overall_result}
    \footnotesize
    \setlength{\tabcolsep}{3pt}
    \begin{tabular}{l|ccccc|ccccc|ccccc|ccccc}
        \bhline{1pt}
        & \multicolumn{10}{c|}{Bipedal robot} & \multicolumn{10}{c}{Quadrupedal robot} \\
        & \multicolumn{5}{c|}{CLAP$_{\mathrm{A}}\!\uparrow$} & \multicolumn{5}{c|}{SAJ$\uparrow$}
        & \multicolumn{5}{c|}{CLAP$_{\mathrm{A}}\!\uparrow$} & \multicolumn{5}{c}{SAJ$\uparrow$} \\
        SNR [dB] & $-10$ & $-5$ & 0 & 5 & 10 & $-10$ & $-5$ & 0 & 5 & 10
        & $-10$ & $-5$ & 0 & 5 & 10 & $-10$ & $-5$ & 0 & 5 & 10 \\
        \hline
        No Processing
        & 0.29 & 0.39 & 0.52 & 0.64 & 0.74 & 1.44 & 1.38 & 1.38 & 1.40 & 1.45
        & 0.16 & 0.26 & 0.40 & 0.55 & 0.68 & 1.76 & 1.68 & 1.64 & 1.65 & 1.71 \\
        SB-INMF~\cite{SBINMF}
        & 0.29 & 0.39 & 0.52 & 0.64 & 0.75 & 1.79 & 1.74 & 1.73 & 1.76 & 1.88
        & 0.22 & 0.31 & 0.43 & 0.57 & 0.68 & 2.07 & 2.00 & 1.96 & 2.06 & 2.28 \\
        Harmonic NMF~\cite{MotorDictionaryNMF}
        & 0.34 & 0.42 & 0.50 & 0.58 & 0.65 & 2.94 & 3.11 & 3.29 & 3.48 & 3.59
        & 0.31 & 0.37 & 0.45 & 0.53 & 0.60 & 2.83 & 2.85 & 2.98 & 3.12 & 3.31 \\
        CLAPSep~\cite{clapsep}
        & 0.32 & 0.42 & 0.63 & 0.71 & 0.75 & 1.88 & 1.97 & 2.92 & 3.08 & 2.83
        & 0.28 & 0.39 & 0.62 & 0.71 & 0.75 & 2.30 & 2.31 & 3.29 & 3.51 & 3.27 \\
        SAM-Audio~\cite{sam_audio}
        & 0.37 & 0.47 & 0.53 & 0.49 & 0.49 & 2.63 & 2.72 & 2.80 & 2.48 & 2.22
        & 0.31 & 0.43 & 0.49 & 0.52 & 0.54 & 2.53 & 2.82 & 3.03 & 2.89 & 2.75 \\
        Conv-TasNet~\cite{ConvTasNet}
        & 0.42 & 0.54 & 0.67 & 0.78 & 0.84 & 2.95 & 3.15 & 3.25 & 3.28 & 3.29
        & 0.39 & 0.50 & 0.64 & 0.75 & \textbf{0.83} & 2.94 & 3.28 & 3.44 & 3.52 & 3.47 \\
        Dual-Path RNN~\cite{DPRNN}
        & 0.33 & 0.44 & 0.52 & 0.58 & 0.61 & 3.15 & 3.37 & 3.49 & 3.57 & 3.61
        & 0.26 & 0.32 & 0.39 & 0.44 & 0.47 & 2.70 & 2.84 & 2.99 & 3.09 & 3.18 \\
        Conformer~\cite{Conformer}
        & 0.45 & 0.59 & 0.70 & 0.79 & \textbf{0.85} & 3.06 & 3.31 & 3.42 & 3.47 & 3.38
        & 0.39 & 0.51 & 0.63 & 0.74 & 0.81 & 2.89 & 3.17 & 3.36 & 3.43 & 3.40 \\
        Transfer-DiT (Ours)
        & \textbf{0.52} & \textbf{0.64} & \textbf{0.73} & \textbf{0.81} & \textbf{0.85} & \textbf{3.29} & \textbf{3.64} & \textbf{3.77} & \textbf{3.78} & \textbf{3.66}
        & \textbf{0.48} & \textbf{0.60} & \textbf{0.70} & \textbf{0.78} & \textbf{0.83} & \textbf{3.22} & \textbf{3.63} & \textbf{3.68} & \textbf{3.72} & \textbf{3.64} \\
        \bhline{1pt}
    \end{tabular}
\end{table*}

\begin{table*}[t]
    \centering
    \caption{
        Classification performance on the evaluation data.
        $\mathrm{AUPRC}_{\mathrm{anom}}$ measures binary discrimination between normal and anomalous sounds, whereas $\mathrm{Acc}_{52}$ measures classification performance among the 52 specific environmental-sound classes.
        The best value in each column is shown in bold.
    }
    \label{tab:classification_result}
    \footnotesize
    \setlength{\tabcolsep}{3pt}
    \begin{tabular}{l|ccccc|ccccc|ccccc|ccccc}
        \bhline{1pt}
        & \multicolumn{10}{c|}{Bipedal robot} & \multicolumn{10}{c}{Quadrupedal robot} \\
        & \multicolumn{5}{c|}{$\mathrm{AUPRC}_{\mathrm{anom}}\!\uparrow$} & \multicolumn{5}{c|}{$\mathrm{Acc}_{52}\!\uparrow$}
        & \multicolumn{5}{c|}{$\mathrm{AUPRC}_{\mathrm{anom}}\!\uparrow$} & \multicolumn{5}{c}{$\mathrm{Acc}_{52}\!\uparrow$} \\
        SNR [dB] & $-10$ & $-5$ & 0 & 5 & 10 & $-10$ & $-5$ & 0 & 5 & 10
        & $-10$ & $-5$ & 0 & 5 & 10 & $-10$ & $-5$ & 0 & 5 & 10 \\
        \hline
        No Processing
        & 0.62 & 0.79 & 0.86 & 0.89 & 0.92 & 0.23 & 0.41 & 0.56 & 0.72 & 0.79
        & 0.61 & 0.81 & 0.84 & 0.90 & 0.92 & 0.11 & 0.27 & 0.48 & 0.63 & 0.70 \\
        SB-INMF~\cite{SBINMF}
        & 0.55 & 0.77 & 0.83 & \textbf{0.91} & 0.90 & 0.14 & 0.36 & 0.48 & 0.61 & 0.71
        & 0.63 & 0.84 & 0.90 & 0.89 & 0.92 & 0.13 & 0.28 & 0.46 & 0.63 & 0.74 \\
        Harmonic NMF~\cite{MotorDictionaryNMF}
        & 0.42 & 0.56 & 0.63 & 0.70 & 0.78 & 0.14 & 0.32 & 0.46 & 0.54 & 0.62
        & 0.42 & 0.49 & 0.64 & 0.74 & 0.81 & 0.11 & 0.23 & 0.34 & 0.45 & 0.55 \\
        CLAPSep~\cite{clapsep}
        & 0.53 & 0.65 & 0.75 & 0.77 & 0.79 & 0.23 & 0.33 & 0.61 & 0.73 & 0.76
        & 0.68 & 0.80 & \textbf{0.92} & 0.93 & 0.91 & 0.17 & 0.34 & 0.58 & 0.70 & 0.76 \\
        SAM-Audio~\cite{sam_audio}
        & 0.51 & 0.65 & 0.63 & 0.43 & 0.52 & 0.23 & 0.36 & 0.47 & 0.38 & 0.37
        & 0.60 & 0.75 & 0.74 & 0.65 & 0.77 & 0.18 & 0.36 & 0.39 & 0.43 & 0.44 \\
        Conv-TasNet~\cite{ConvTasNet}
        & 0.57 & 0.81 & 0.89 & 0.90 & 0.92 & 0.20 & 0.41 & 0.62 & 0.72 & 0.79
        & 0.53 & 0.79 & 0.90 & 0.91 & 0.94 & 0.14 & 0.34 & 0.56 & 0.70 & 0.78 \\
        Dual-Path RNN~\cite{DPRNN}
        & 0.51 & 0.65 & 0.70 & 0.77 & 0.81 & 0.17 & 0.28 & 0.44 & 0.52 & 0.56
        & 0.30 & 0.41 & 0.58 & 0.63 & 0.70 & 0.06 & 0.12 & 0.19 & 0.27 & 0.33 \\
        Conformer~\cite{Conformer}
        & 0.67 & \textbf{0.85} & 0.87 & 0.90 & 0.93 & 0.25 & 0.51 & 0.71 & \textbf{0.81} & \textbf{0.86}
        & 0.57 & 0.80 & 0.85 & 0.92 & 0.94 & 0.17 & 0.38 & 0.59 & 0.71 & 0.82 \\
        Transfer-DiT (Ours)
        & \textbf{0.71} & 0.73 & \textbf{0.93} & 0.89 & \textbf{0.94} & \textbf{0.43} & \textbf{0.62} & \textbf{0.76} & \textbf{0.81} & \textbf{0.86}
        & \textbf{0.77} & \textbf{0.88} & 0.91 & \textbf{0.97} & \textbf{0.97} & \textbf{0.34} & \textbf{0.55} & \textbf{0.71} & \textbf{0.81} & \textbf{0.86} \\
        \bhline{1pt}
    \end{tabular}
\end{table*}

Figure~\ref{fig:examples} shows examples of ego-noise separation results.
For the bipedal robot, the left side of \cref{fig:examples} shows that Conformer~\cite{Conformer} excessively suppresses the environmental sound when the ego-noise is strong.
The proposed method preserves the environmental sound during the same intervals.
For the quadrupedal robot, the right side of \cref{fig:examples} shows residual ego-noise with Conformer.
The proposed method suppresses the ego-noise more effectively.

\Cref{tab:training_strategy_comparison} compares two training strategies: training the DiT from scratch and transfer learning.
For both robots, transfer learning stopped substantially earlier than training from scratch.
Transfer learning also achieved considerably higher CLAP$_{\mathrm{A}}$, SAJ, $\mathrm{Acc}_{52}$, and $\mathrm{AUPRC}_{\mathrm{anom}}$ values.
The faster convergence and higher metric values support the use of pretrained-model transfer for robot-specific ego-noise separation.

\Cref{tab:overall_result} summarizes the ego-noise separation performance, while \cref{tab:classification_result} summarizes the downstream ESC performance. 
Without any processing, $\mathrm{AUPRC}_{\mathrm{anom}}$ decreased particularly under the low-SNR condition. 
The no-processing baseline retained relatively high $\mathrm{AUPRC}_{\mathrm{anom}}$ because the anomalous sound classes contain acoustically salient events.
Even so, the degradation at low SNRs confirms that legged-robot ego-noise interferes with ESC. 
Without processing, the CLAP$_{\mathrm{A}}$ scores were also low, especially for the quadrupedal robot.
The degradation in both metrics shows the importance of ego-noise separation when footsteps and actuator sounds dominate the recording.

Without processing, the CLAP$_{\mathrm{A}}$ scores were consistently lower for the quadrupedal robot than for the bipedal robot at all SNR levels.
The corresponding $\mathrm{AUPRC}_{\mathrm{anom}}$ values were similar between the two robots.
The lower CLAP$_{\mathrm{A}}$ scores indicate greater degradation of the environmental-sound content by the dense footstep impacts and motor noise of the quadrupedal robot.
The acoustically salient anomalous events nevertheless remained detectable by the classifier.
Transfer-DiT improved $\mathrm{AUPRC}_{\mathrm{anom}}$ relative to no processing in all quadrupedal conditions, whereas the effect varied across bipedal conditions.
The CLAP$_{\mathrm{A}}$ improvement over no processing was larger for the quadrupedal robot at all SNR levels.

% The proposed method was effective for both platforms despite the contrasting ego-noise structures shown in \cref{fig:ego_noise_characteristics}.
% The source-separation models used outdoor and lounge recordings for pre-deployment adaptation.
% The evaluation used separately recorded lobby and corridor audio.
% The results therefore assess robustness to combined changes in floor material, room acoustics, and gait.
% The evaluation does not isolate floor-material effects or establish generalization to every possible floor structure.

The zero-shot neural separators showed limited and unstable effectiveness. 
CLAPSep~\cite{clapsep} improved the CLAP$_{\mathrm{A}}$ and SAJ scores in several conditions.
The improvements indicate that text-prompt-based separation can remove some ego-noise components. 
SAM-Audio also improved SAJ scores compared with no processing, but $\mathrm{AUPRC}_{\mathrm{anom}}$ was often lower than that of the no-processing baseline, especially for the bipedal robot. 
The inconsistent gains show that directly applying general-purpose zero-shot neural separators to legged security robots does not reliably improve ESC performance.

The supervised neural separators trained using the clips selected by RecurGraph showed architecture-dependent performance.
Conv-TasNet~\cite{ConvTasNet} and Conformer~\cite{Conformer} generally improved separation quality and maintained or improved downstream performance in many conditions.
Dual-Path RNN~\cite{DPRNN} improved SAJ but often reduced CLAP$_{\mathrm{A}}$ and $\mathrm{AUPRC}_{\mathrm{anom}}$ relative to no processing.
These results show that the selected clips can support robot-specific supervised training, although downstream performance depends strongly on the separator architecture.
The proposed Transfer-DiT achieved the best or competitive ESC performance across all robot and SNR conditions. 
The improvement over no processing was particularly clear in the most challenging low-SNR condition. 

% For the quadrupedal robot, Transfer-DiT achieved the highest $\mathrm{AUPRC}_{\mathrm{anom}}$ at all SNR levels. 
% For the bipedal robot, Transfer-DiT was tied with the best supervised neural separator in one condition.
% The difference among the top-performing separators was small in the easiest high-SNR condition because the no-processing baseline already achieved high $\mathrm{AUPRC}_{\mathrm{anom}}$. 
% The largest gains occurred in difficult acoustic conditions, where robot ego-noise strongly masks anomalous sounds.

% In terms of separation quality, Transfer-DiT achieved the highest SAJ score in all robot and SNR conditions. 
% Transfer-DiT also achieved the highest or competitive CLAP$_{\mathrm{A}}$ scores. 
% At 10~dB for the bipedal robot, Conformer and Transfer-DiT achieved the same CLAP$_{\mathrm{A}}$ score.
Transfer-DiT maintained strong CLAP$_{\mathrm{A}}$ scores, achieved the best SAJ scores, and showed strong $\mathrm{AUPRC}_{\mathrm{anom}}$ performance.
The consistent performance across these metrics indicates that Transfer-DiT preserves environmental sounds while suppressing robot ego-noise.

\subsection{Physical Playback Evaluation}

\begin{table}[t]
    \centering
    \caption{
        Separation and classification performance on the physical playback recording with the walking quadrupedal robot.
        CLAP$_{\mathrm{T}}$ and SAJ are reported as means over 14 clips; $\mathrm{AUPRC}_{\mathrm{anom}}$ and $\mathrm{Acc}_{52}$ are computed over the same clips.
        The highest value in each column is shown in bold.
    }
    \label{tab:physical_playback_result}
    \footnotesize
    \setlength{\tabcolsep}{2pt}
    \begin{tabular}{l|cccc}
        \bhline{1pt}
        Method
        & CLAP$_{\mathrm{T}}\!\uparrow$
        & SAJ$\uparrow$
        & $\mathrm{AUPRC}_{\mathrm{anom}}\!\uparrow$
        & $\mathrm{Acc}_{52}\!\uparrow$ \\
        \hline
        No Processing & 0.15 & 1.23 & \textbf{1.00} & 0.43 \\
        SB-INMF~\cite{SBINMF} & 0.15 & 1.37 & \textbf{1.00} & 0.36 \\
        CLAPSep~\cite{clapsep} & 0.17 & 1.39 & \textbf{1.00} & 0.57 \\
        SAM-Audio~\cite{sam_audio} & \textbf{0.25} & 2.28 & \textbf{1.00} & 0.64 \\
        Conv-TasNet~\cite{ConvTasNet} & 0.14 & 1.71 & 0.94 & 0.14 \\
        Dual-Path RNN~\cite{DPRNN} & 0.05 & 2.05 & 0.76 & 0.29 \\
        Conformer~\cite{Conformer} & 0.13 & 1.58 & \textbf{1.00} & 0.57 \\
        Transfer-DiT (Ours) & \textbf{0.25} & \textbf{2.45} & \textbf{1.00} & \textbf{0.71} \\
        \bhline{1pt}
    \end{tabular}
\end{table}

Audio and video examples from the physical playback experiment are available on the \projectpage.
\cref{tab:physical_playback_result} summarizes the separation and classification results for the physical playback recording.
All applicable separators increased the mean SAJ score relative to No Processing.
The increase indicates that ego-noise separation generally improved the predicted perceptual quality of the simultaneously recorded playback sounds according to SAJ.
Transfer-DiT tied with SAM-Audio for the highest mean CLAP$_{\mathrm{T}}$ score and achieved the highest mean SAJ score and $\mathrm{Acc}_{52}$.
SAM-Audio achieved the second-highest mean SAJ score and $\mathrm{Acc}_{52}$.

$\mathrm{Acc}_{52}$ reveals differences in fine-grained event recognition.
This ranking differs from that in \cref{tab:overall_result}, where the supervised neural separators generally outperformed SAM-Audio.
The physical playback evaluation differs from the controlled reference-audio evaluation in the recording process and the source-specific acoustic transfer paths realized during walking.
The physical playback recording was also collected at a different site.
Site-dependent room acoustics and floor conditions may therefore have contributed to the ranking difference.
SAM-Audio is a general-purpose zero-shot neural separator and does not rely on adaptation to robot-specific training mixtures.
SAM-Audio's relatively strong performance is consistent with a potential benefit of its general-purpose separation prior under the physical recording conditions.

% Because these factors changed together, the physical playback experiment cannot identify the contribution of each factor.
% The two protocols instead provide complementary evidence.
% The reference-audio evaluation supports controlled comparisons across SNRs and robot platforms using held-out-site recordings.
% The physical playback evaluation examines end-to-end separator behavior through the physical recording chain at a different site.

Transfer-DiT tied for the highest CLAP$_{\mathrm{T}}$ and $\mathrm{AUPRC}_{\mathrm{anom}}$ and achieved the highest SAJ and $\mathrm{Acc}_{52}$.
Transfer-DiT remained effective across the systematically varied held-out-site conditions in the controlled reference-audio evaluation.
Transfer-DiT also remained effective through the physical recording chain at a previously unseen site.

Future work will investigate periodic offline adaptation using continual-learning techniques~\cite{continual_learning_review}.
To improve robustness across the target operating domain, adaptation recordings should cover as many representative sites and routes as practical.
Recordings acquired during deployment could be retained for subsequent adaptation cycles between deployment periods.
Clips associated with detected anomalous events would be excluded before adaptation.
Continual-learning techniques would be used to incorporate newly encountered acoustic conditions while mitigating catastrophic forgetting of conditions represented in earlier adaptation data.

\section{Conclusion}
To achieve the objective of obtaining robot- and domain-specific supervision from unlabeled pre-deployment recordings, this paper proposed an open-set ego-noise separation framework for legged-robot audition that combines annotation-free adaptation with pretrained-model transfer.
RecurGraph automatically selects ego-noise-dominant clips from unlabeled pre-deployment recordings.
The selected clips are mixed with diverse environmental sounds sampled from a large-scale sound-event dataset to provide paired mixture--target supervision for open-set ego-noise separation.
Transfer-DiT adapts a pretrained general-purpose separator to realize this removal function with high fidelity.
Experiments with bipedal and quadrupedal robots showed that RecurGraph reliably selected ego-noise-dominant clips and that Transfer-DiT improved separation quality and downstream classification performance compared with baseline separators.
The clip-selection and separation results demonstrate that the proposed framework achieved the stated objective through annotation-free adaptation.

\bibliographystyle{IEEEtranStyle_LimitNumAuthors}
\bibliography{1-reference}

\end{document}